\documentclass{article} 
\usepackage[vmargin=1.18in,hmargin=1.1in]{geometry}
\usepackage{amsmath,amssymb,amsfonts,graphicx,amsthm,nicefrac,mathtools,bm}
\usepackage{hyperref} 
\usepackage[numbers]{natbib}
\usepackage[english]{babel}
\usepackage{times}
\usepackage[T1]{fontenc}
\usepackage{bbm,mdframed}
\usepackage[dvipsnames]{xcolor}
\usepackage{xspace}
\usepackage{rotating,multirow} 
\usepackage{tikz}
\usetikzlibrary{arrows}
\usetikzlibrary{shapes} 
\usepackage{algorithm,algorithmic} 
\usepackage{fancyhdr} 
\usepackage{tabularx}
\usepackage[utf8]{inputenc}
\usepackage[english]{babel}
\usepackage{amsthm}
\newtheorem{lemma}{Lemma}
\usepackage{booktabs} 
\usepackage{amsmath,amssymb,latexsym,graphicx,epsfig,psfrag,float, multicol} 
\usepackage{paralist} 
\usepackage{graphicx,subfig}
\usepackage{graphics}
\usepackage{tabularx}
\usepackage{appendix}

\newcommand{\R}{\mathbb{R}}

\newcommand{\bw}{{\boldsymbol{w}}}
\newcommand{\bx}{{\boldsymbol{x}}}

\newcommand{\bv}{{\boldsymbol{v}}}

\date{}
\author{\theauthor}
\title{\thetitle} 
\fancypagestyle{plain}{%
  \fancyhf{}%
  \lhead{\thepage}
}

\newcommand{\theauthor}{}
\newcommand{\thetitle}{History PCA: A New Algorithm for Streaming PCA}

\newtheorem{theorem}{Theorem}
\begin{document}

\author{ 
Puyudi Yang,\hspace{1mm} Cho-Jui Hsieh,\hspace{1mm} Jane-Ling Wang \\
University of California, Davis\\
\texttt{pydyang, chohsieh, janelwang@ucdavis.edu}
} 
\maketitle
\vspace{-5mm}

\begin{abstract} 
In this paper we propose a new algorithm for streaming principal component analysis. With limited memory, small devices cannot store all the samples in the high-dimensional regime. Streaming principal component analysis aims to find the $k$-dimensional subspace which can explain the most variation of the $d$-dimensional data points that come into memory sequentially. In order to deal with large $d$ and large $N$ (number of samples), most streaming PCA algorithms update the current model using only  the incoming sample and then dump the information right away to save memory. However the information contained in previously streamed data could be useful. Motivated by this idea, we develop a new streaming PCA algorithm called History PCA that achieves this goal. By using $O(Bd)$ memory with $B\approx 10$ being the block size, our algorithm converges much faster than existing streaming PCA algorithms. By changing the number of inner iterations, the memory usage can be further reduced to $O(d)$ while maintaining a comparable convergence speed. We provide theoretical guarantees for the convergence of our algorithm along with the rate of convergence. We also demonstrate on synthetic and real world data sets that our algorithm compares favorably with other state-of-the-art streaming PCA methods in terms of the convergence speed and performance. 
\end{abstract}

\section{Introduction}   

\setlength{\abovedisplayskip}{3pt}
\setlength{\abovedisplayshortskip}{1pt}
\setlength{\belowdisplayskip}{3pt}
\setlength{\belowdisplayshortskip}{1pt}
\setlength{\jot}{3pt}
\setlength{\textfloatsep}{3pt}  

Principal component analysis (PCA) is one of the most fundamental tools for feature extraction in machine learning problems such as data compression, classification, clustering, image processing and visualization.


Given a data set $X\in \R^{N\times d}$ with $N$ samples and $d$ features, 
PCA can be easily
 performed through eigen decomposition of the sample covariance
matrix $\frac{1}{N} X^\top X$ or the singular value decomposition of the data matrix $X$, and the theoretical
guarantee is established by matrix Bernstein theory \cite{RV10a, tropp12}.  Throughout the paper  we assume that $X$ has mean zero for notation simplicity. 
For X
with large $N$ and $d$, we may not be able to store matrices of size $N\times d$ or $d\times d$ due to limited memories on small devices such as laptops and mobile phones, and it is often expensive to conduct multiple passes over the data. 
For these cases, streaming PCA algorithms \cite{IM13a,MH14a,CDS15a,EO82a,PJ16a,EL13, CB15a, CB15b, MG15a, MG15b} come into play, with the nice property that they only need to use $O(d)$ or $O(Bd)$ memory to store the current sample or the current block of samples. 

While  these streaming PCA algorithms resolve the memory issues, 
they often suffer from slow convergence in practice. 
The main reason is that most streaming PCA algorithms in the past decades do not fully utilize  information in previously streamed data, which came in sequentially and contain valuable information. 
 This motivates us to develop a new algorithm that utilizes past information effectively without taking much memory space. 

Specifically, our new algorithm can compute the top-$k$ principal components in the streaming
setting with $O(Bd)$ memory requirement,  where $B$ is the block size and can be pretty small (e.g., $1$ or $10$). 
The convergence speed of the new algorithm is much faster than existing approaches mainly due to effective use of past information, and by setting the number of inner iterations to one, 
the memory cost of our algorithm can be reduced to $O(d)$.   The resulting algorithm is similar
to Oja's algorithm but with the ability to compute the top-$k$ principal components altogether without the need to tune the step size as in Oja's algorithm.
We provide theoretical justification of our algorithm. 

Experimental results show that our algorithm consistently outperforms existing algorithms in the streaming setting. 
Moreover,  we test the ability of our algorithm to compute PCA on massive data sets. Surprisingly it outperforms existing streaming and non-streaming algorithms (including VR-PCA and power method)
in terms of both the number of data passes and real run time.

The rest of the paper is outlined as follows. First, we discuss and compare current algorithms on streaming PCA. Then, we propose the History PCA algorithm with theoretical guarantees. Finally, we provide experiments on both simulated streaming data sets and real world data sets.

\section{Related Work}
\label{sec:related}

There are many existing methods to  estimate the top $k$ eigenvectors of the true covariance matrix.
It was shown by Adamczak et al. \cite{RA10a} that the sample covariance matrix can guarantee estimation of the true covariance matrix with a fixed error in the operator norm for sub-exponential distributions, if the sample size is $O(d)$. Later, this statement is generalized to  distributions with finite fourth moment and the true eigenvectors of the distribution can be recovered with high probability by the singular value decomposition of the sample covariance matrix formed by $O(d)$ samples \cite{RV10a}.

\subsection{Non-streaming PCA Algorithms}

When the full data set is given in advance, the optimal way to estimate principal components
is to conduct eigen decomposition on the sample covariance matrix $\frac{1}{N} X^\top X$. 
However, it is often impossible to form the $d \times d$ sample covariance matrix for large-scale data sets. 
So we need to compute $\frac{1}{N}(X^\top X) v=\frac{1}{N} X^\top (X v)$ in the power method without explicitly forming the sample 
covariance matrix. 

Recently, Shamir \cite{OS15a, OS16a} proposed the VR-PCA method that outperforms power method for PCA computation. 
The main idea is to reformulate PCA as a stochastic optimization problem and apply the variance reduction techniques~\cite{RJ13a}.  
Although VR-PCA looks similar to streaming PCA, 
it cannot obtain the first update until the second pass. So it is not suitable in the streaming setting. 


Since the entire data are often too large to be stored on the hard disk, algorithms that are specifically designed for  streaming data have  become a research focus in the past few decades.


\subsection{Streaming PCA Algorithms}

A classical algorithm for streaming PCA is Oja's algorithm \cite{EO82a}. 
It can be viewed as the traditional stochastic gradient decent method, in which each data point that come into memory can be viewed as a random sample drawn from an underlying distribution. In 2013, Balsubramani et al. \cite{AB13a} modified Oja's algorithm with a special form of step size and provided the statistical analysis of this algorithm. 
Later on in 2015, Jin et al. \cite{CJ15a} proposed a faster algorithm to find the top eigenvector based on shift and invert framework with improved sample complexity. Although this algorithm seem to have good sample complexity, it requires the initialization vector be within certain distance to the true top eigenvector, which is costly to attain and hard to achieve when the dimension $d$ is very high.
More recently, Jain et al. \cite{PJ16a} proved that Oja's algorithm can achieve the optimal sample complexity 
for the first eigenvector under certain choices of step size, 
in the sense that it matches the matrix Bernstein inequality \cite{RV10a, tropp12}. 
However, the step sizes depend on some data-related constants that cannot be estimated beforehand. So in practice one still needs to test different step sizes and choose the best one. 
%
%
Under a similar assumption on the step size, Li et al. \cite{CL17a} recently proved that the streaming PCA algorithm matches the minimax information rate, and Allen-Zhu and Li \cite{ZA17a} extended such optimality to the k-subspace version and proposed the Oja$^{++}$ algorithm.
Recently, Sa et al. \cite{CDS15a} showed the global convergence of Alecton algorithm, which is another variant of stochastic gradient descent algorithm applied to a non-convex low-rank factorized problem. In our experiments, we compare our algorithms with Oja's algorithm and Oja$^{++}$ algorithm based on various choices of step sizes and show that our algorithm outperforms all of them.

Mitliagkas et al. \cite{IM13a} introduced the block stochastic power method. This method uses each block of data to construct an estimator of the covariance matrix and applies power method in each iteration to update eigenvectors. Theoretical guarantees are provided under the setting of the spiked data model. More recently, Li et al. \cite{CL15a} extended the work of \cite{IM13a} by using blocks of variable size and progressively increasing the block size as the estimate becomes more accurate. 
Later on, Hardt and Price \cite{MH14a} gave a robust convergence analysis of a more general version of block stochastic power method for arbitrary distributions.  

The aforementioned works are all in the setting of streaming i.i.d. data samples. There is another line of research for an  arbitrary sequence of data. Most works  \cite{EL13, CB15a, CB15b, MG15a, MG15b} use the technique of computing a sketch of matrix to find an estimator of the top eigenvector.




Our algorithm is under the setting of streaming i.i.d. data samples and it differs from other streaming algorithms in the sense that our History PCA algorithm exploits key  summaries in the historical data to achieve better performance. 
 


\section{Problem Setting and Background}
\label{sec:background}

Let $\bx_1, \bx_2, ..., \bx_N$ be a sequence of $d$-dimensional data that stream into memory. Assume all the $x_i$'s are sampled i.i.d from a distribution with zero-mean and a fixed covariance matrix $\Sigma \in \R^{d \times d}$. Let $\lambda_1 > \lambda_2 \geq \lambda_3 \cdots \geq \lambda_d$ and $\bv_1, \cdots, \bv_d$ denote the eigenvalues and corresponding eigenvectors of $\Sigma$. 


In this paper, our goal is to recover the top $k$ eigenvectors. In the case of $k = 1$, our goal is to find a unit vector that is within the $\varepsilon$-neighborhood of the first true eigenvector $v_1$. In the general case with $k > 1$, we use the the largest-principal-angle-based distance function as the metric to evaluate our algorithm, which is 
\begin{align}
d(span(U), span(V)) &= d(U, V) 
= \| U_{\bot}^\top V \|_2 = \|V_{\bot}^\top U\|_2 \label{eq:principal_angle}
\end{align}
for any $U, V \in \R^{d \times k}$. Here $U_{\bot}$ denotes the $d \times (d-k)$ orthogonal basis of the perpendicular subspace to the subspace spanned by $d \times k$ matrix $U$.

Our algorithm is motivated by the block stochastic power method \cite{IM13a}. While the classical power method keeps on multiplying the vector with a fixed sample covariance matrix at each iteration ($\bw_{i} = \frac{1}{N} X^\top X\bw_{i-1}$), the block stochastic power method \cite{IM13a} updates the current estimate by 
\begin{align}
    \bw_{i} &\leftarrow \frac{1}{B} X_i^\top X_i \bw_{i-1}
\end{align}
where $\frac{1}{B} X_i^\top X_i$ is 
a sample covariance matrix formed by the block of $B$ data points that stream into memory at iteration $i$. Thus, the block stochastic power method can be viewed as a stochastic variant of the power method, 
and using a larger block size at each iteration can reduce the 
potentially large variance by averaging out the noises. 

The block stochastic power method has several limitations.   First, the block size $B$ needs to be very large in order for this algorithm to work. That is mainly because the algorithm requires the sample covariance matrix formed by each block of data points to be very close to the true covariance matrix, which is difficult to  achieve if the block size $B$ is not large enough. And this block size $B$ depends highly on the target accuracy $\varepsilon$. Secondly, the block stochastic power method cannot converge to the true eigenvectors of $\Sigma$. If the block size $B$ is fixed, it can only converge to the $\varepsilon$-approximated solution, where $\varepsilon=O(1/B)$. 

Our algorithm aims at solving the above problems by using more past information. Instead of doing only one matrix-vector multiplication at each step after a block of data stream in, we do matrix-vector multiplication iteratively either until the algorithm converges or for a pre-specified number of times. This strategy gives us good results at very early stage.    
Moreover, with the use of past information, we could further reduce the variance of the estimates. Since intuitively there is no reason to favor a particular past block, we assign equal weights to all data samples (or blocks) in each iteration.  Details are provided below. 

\section{Proposed Algorithms}
\label{sec:proposed}

Since data come into memory sequentially, we can form  blocks of sample, each  with size $B$, as data streaming in, and denote the $i$th data block by a $B \times d$ matrix $X_i = (\bx_{i1}, \bx_{i2}, \ldots, \bx_{iB})^T$, where $x_{it}$ is a data point of dimension $d$ for $t = 1, \cdots, B$.

\subsection{Main algorithm: History PCA (Rank-$1$)}

We first provide the method and intuition of our History PCA in the rank-$1$ case.
 
At time $1$, we have the first block of data $X_1$. Since we have no past information, we can make use of the sample covariance matrix formed by the first block of data point, $\frac{1}{B} X_1^\top  X_1 = \frac{1}{B} \sum_{t = 1}^{B} \bx_{1t} \bx_{1t}^\top $, to estimate the true covariance matrix and its eigenvectors. But instead of finding the eigenvectors of $\frac{1}{B} X_1^\top  X_1$, we try to find the eigenvectors of $(I+\frac{1}{B} X_1^\top  X_1)$ in the first step, which is mainly for our convergence theory to go through. In the rank-$1$ case, we only need to save 
the first eigenvector $\bw_{1}$ of $(I+\frac{1}{B} X_1^\top  X_1)$.

At time $2$, we have the second block of data $X_2$. If we could have both blocks of data in memory, then the best we can do is to form the sample covariance matrix $\frac{1}{2}( \frac{1}{B} X_2^\top  X_2 + \frac{1}{B} X_1^\top  X_1)$. However, limited memory space prohibits us from  storing the entire first block of data. Thus, we need to find an alternative scheme to extract key information from the first block so that the estimator at time $2$ could be better than just using the sample covariance matrix $\frac{1}{B} X_2^\top  X_2$. 
For simplicity of illustration we assume $\lambda_1 = 1$.  An intuitive idea is to form a rank-$1$ matrix $\lambda_1 \bw_{1} \bw_{1}^\top = \bw_{1} \bw_{1}^\top$ and make use of it in our estimator. Intuitively, we should assign equal weights to the information we get from both blocks of data. Thus, the updated estimator is  $\frac{1}{2}( \frac{1}{B} X_2^\top  X_2 +  \bw_{1} \bw_{1}^\top )$. 
Now we only need to save the first eigenvector $\bw_{2}$ of this estimator. 

At time $\tau$, we have the $\tau$th block of data $X_\tau$. Similarly, we could get a new estimator of the covariance matrix by exploiting the history from the past $\tau-1$ blocks of data. Intuitively, each block of data should have equal weights. Thus, we assign the weight $\frac{1}{\tau}$ and $\frac{\tau-1}{\tau}$, respectively, to the sample covariance matrix formed by $X_\tau$ and the rank-$1$ matrix formed by the past $\tau-1$ blocks of data. Therefore, the new covariance estimator at time $\tau$ is
\begin{equation}
    \frac{1}{\tau}\frac{1}{B} X_{\tau}^\top X_{\tau} + \frac{\tau-1}{\tau} \bw_{\tau-1} \bw_{\tau-1}^\top, 
    \label{eq:tau}
\end{equation}
where $\bw_{\tau-1}$ is the eigenvector at time $\tau-1$. 

At the final time $n$, we have seen in total $n$ blocks of data and we output $\bw_{n}$.
We summarize this algorithm in Algorithm~\ref{alg:main_k1}. 


\begin{algorithm}[t]
\caption{History PCA: k = 1 \label{alg:main_k1}}
\begin{algorithmic}
    \STATE  Input: $\{X_1, \ldots, X_n \}$, block size: $B$. 
    \STATE $\bw_0 \sim N(0, I_{d \times d})$. $\bw_1 \gets \bw_0 /\|\bw_0 \|_2$
    \WHILE{ $\bw_1$ not converge }  
        \STATE $\bw_1 \gets \bw_1 + \frac{1}{B} X_{1}^\top X_{1} \bw_1$ 
        \STATE $\bw_1 \gets \bw_1 /  \| \bw_1 \|_2$
    \ENDWHILE 
    
    \FOR{$\tau = 2, \ldots, n$} 
        \STATE $ \bw_{\tau} = \bw_{\tau - 1}$
        \WHILE{ $\bw_{\tau}$ not converge }
            \STATE $\bw_{\tau} \gets \frac{\tau - 1}{\tau}   \bw_ {\tau - 1} \bw_ {\tau - 1}^\top \bw_ {\tau} +  \frac{1}{\tau} \frac{1}{B} X_{\tau}^\top X_{\tau}   \bw_{\tau}$         
            \STATE $\bw_{\tau} \gets \bw_{\tau} /  \| \bw_{\tau} \|_2$
        \ENDWHILE
    \ENDFOR
    \STATE  Output $\bw_n$    
    \STATE Extension: the "while" loops can be run by $m$ iterations. 
\end{algorithmic}  
\end{algorithm}
\vspace{-0.5mm}
\subsection{Solving each subproblem approximately}

In Algorithm~\ref{alg:main_k1}, at each iteration $\tau$ we run the power method on the current
covariance estimator, which is equation~(\ref{eq:tau}), to get the first eigenvector. This can be viewed as a subproblem. However, it will in theory 
require infinite number of iterations to get the exact eigenvector. Simulations show that using exact convergence have similar results as using power method with small $m$ iterations. So instead of solving the sub-problem exactly, we solve it approximately by performing matrix vector multiplication $m$ times in the for loops.
Note that the space complexity of our algorithm is $O(Bd)$. However, when we set $m = 1$ (only one power iteration at each step), Algorithm~\ref{alg:main_k1} will only require $O(d)$ memory. 
This gives us the flexibility
to reduce the memory size to $O(d)$ if memory is not enough. 

\subsection{Extending to rank-$k$ History PCA}

Our method can be generalized to find the first $k$ eigenvectors, as shown in Algorithm~\ref{alg:main_k}. In the rank-$k$ case, we can simply replace the $d$-dimensional vector $\bw$ with a $d \times k$ matrix $Q$ and then replace the normalization update by the QR decomposition. 

In the rank-$k$ case, we make use of the history information of eigenvalues. So the new covariance estimator at time $\tau$, which we utilize to update our estimate, is 
\begin{equation}
       \frac{1}{\tau} \frac{1}{B} X_{\tau}^\top X_{\tau} + \frac{\tau - 1}{\tau} Q_{\tau - 1} \Lambda_{\tau - 1} Q_{\tau - 1}^\top 
      \label{eq:tau_k}
\end{equation}
where $Q_{\tau - 1}$ is a matrix of $k$ eigenvectors and $\Lambda_{\tau - 1}$ is a $k \times k$ diagonal matrix with corresponding eigenvalues as the diagonal elements at time $\tau-1$.

Similarly, in the generalized History PCA algorithm, we can replace the exact solver by an approximate solver (with 1 or a fixed number of iterations $m$) to find $Q_\tau$. We can further reduce the memory complexity of our algorithm from $O((k+B)d)$ to $O(kd)$ when we set $m = 1$.

\begin{algorithm}[h]
\caption{History PCA:  $k \geq 1$ \label{alg:main_k}}
\begin{algorithmic}
    \STATE  Input: $\{X_1, \ldots, X_n \}$, block size: $B$. 
    \STATE $H^i \sim N(0, I_{d \times d}), 1 \leq i \leq k$. $H = Q_1 R_1$ (QR-decomposition)
    \WHILE{ $Q_1$ not converge }
	  \STATE  $S_1 \gets Q_1 +  \frac{1}{B} X_{1}^\top X_{1}   Q_1$ 
	  \STATE $S_1 = Q_1 R_1$ (QR-decomposition)    
    \ENDWHILE
    \STATE $\lambda_j = \| S_1[:j] \|_2$ for $j = 1, \cdots, k$. 
    \STATE $\Lambda_1 = diag (\lambda_1, \cdots, \lambda_k)$    
    \FOR{$\tau = 2, \ldots, n$} 
        \STATE $ Q_{\tau} = Q_{\tau - 1}$
        \WHILE{ $Q_{\tau}$ not converge }  
            \STATE $S_{\tau} \gets  \frac{\tau - 1}{\tau} Q_{\tau - 1} \Lambda_{\tau - 1} Q_{\tau - 1}^\top  Q_{\tau} +  \frac{1}{\tau} \frac{1}{B} X_{\tau}^\top X_{\tau}   Q_{\tau}$
            \STATE $S_{\tau} = Q_{\tau} R_{\tau}$ (QR-decomposition)
        \ENDWHILE 
        \STATE $\lambda_j = \| S_{\tau}[:j] \|_2$ for $j = 1, \cdots, k$. 
        \STATE $\Lambda_{\tau} = diag (\lambda_1, \cdots, \lambda_k)$   
    \ENDFOR
    \STATE  Output $Q_{\tau}$ 
    \STATE Extension: the "while" loops can be run by $m$ iterations. 
\end{algorithmic}  
\end{algorithm}
\subsection{Theoretical Analysis}

Now we provide the convergence theorem for the proposed algorithm. 
Let $A_i = \frac{1}{B} X_i^\top  X_i \in \R^{d \times d}$ be the sample covariance matrix formed by $i$-th block of data, for $i = 1, \cdots, n$, that satisfies the following assumptions:

\begin{compactitem}
\item $E[A_i] = \Sigma$, where $\Sigma \in \R^{d \times d}$ is a symmetric PSD matrix.
\item $\|A_i - \Sigma\|_2 \leq M$ with probability $1$.
\item $\max \{ \|E[(A_i - \Sigma)(A_i - \Sigma)^\top]\|_2, \|E[ (A_i - \Sigma)^\top (A_i - \Sigma)]\|_2\}\leq \mathcal{V}$.
\end{compactitem}

Let $\bv_1, \cdots, \bv_d$ denote the eigenvectors of $\Sigma$ and $\lambda_1 > \lambda_2 \geq \lambda_3 \cdots \geq \lambda_d$ denote the corresponding eigenvalues. Our goal is to compute an $\varepsilon$-approximation to $\bv_1$ which  is a unit vector $\bw$  satisfying $\sin^2(\bw, \bv_1) = 1- (\bw^\top \bv_1)^2 \leq \varepsilon$, where $\sin(\bw, \bv_1)$ denotes the sin of the angle between $\bw$ and $\bv_1$. In the following, we prove that our History PCA algorithm can achieve $\varepsilon$ accuracy with 
\begin{equation}
\varepsilon = O(\frac{\mathcal{V}}{|2(\lambda_1 - \lambda_2) - 1|} \frac{1}{n})
\label{eq:aaa}
\end{equation}
after seeing $n$ data blocks.

 Jain et al. \cite{PJ16a} proved the convergence of their algorithm by directly analyzing the convergence of Oja's algorithm as an operator on the initialized vector rather than analyzing the error reduced after each update. Oja's algorithm applies the matrix$$B_n = (I+ \eta_n A_n)(I + \eta_{n-1}A_{n-1})\cdots(I+\eta_1A_1)$$ to the random initial vector, say $\bw_0$, and then output the normalized result $$\bw_n = \frac{B_n \bw_0}{\|B_n \bw_0\|_2}.$$ Inspired by their approach, we apply the following trick to prove the theoretical guarantees of our algorithm. If we set $m =1$, our algorithm can also be viewed as applying $B_n$ on the initialized vector $\bw_0$. In our algorithm, we have $\eta_1 = 1$ and $\eta_i = \frac{1}{i - 1}$ for $i > 1$.
 We summarize the main convergence theorem for the History PCA algorithm here. The detailed proof can be found in the appendix. 
 
\begin{theorem}
Fix any $\delta > 0$. Let $n_0$ be the smallest integer such that $n_0 > \max( 4 \mathcal{M}, 18 \frac{\mathcal{V} +\lambda_1^2}{\log (1 + \frac{\delta}{100})}).$ Let $n > n_0$. Under the assumption that $B_n$ is invertible, the History PCA algorithm (rank-$1$ with $m=1$) converges to an $\varepsilon$-accurate solution
with 
\begin{equation}
 \varepsilon_n = O(\frac{\mathcal{V}}{|2(\lambda_1 - \lambda_2) - 1|} \frac{1}{n}),
\end{equation}
with probability at least $1 - \delta$. 
\label{thm: main}
\end{theorem}

To prove our main Theorem~\ref{thm: main}, we need to introduce the following lemma.  
\begin{lemma}
Assume $B_n \in \mathbb{R}^{d \times d}$ is invertible for any positive integer $n$. Let $n_0$ be a non-negative integer. Let $C_{n, n_0}  \in \mathbb{R}^{d \times d}$ be such that $B_{n+n_0} = C_{n, n_0} B_{n_0}$ for any positive integer $n$. Let $\bv \in \mathbb{R}^d$ be a unit vector, and let $V_\perp$ be a matrix whose columns form an orthonormal basis of the subspace orthogonal to $\bv$. If $\bw \in \mathbb{R}^d$ is chosen uniformly at random from the surface of the unit sphere, then with probability at least $1 - \delta$,
$$\sin^2(\bv, \frac{B_{n+n_0}\bw}{\|B_{n+n_0}\bw\|_2})  \leq \frac{c \log (1 / \delta)}{\delta} \frac{  Tr(V_\perp^\top C_{n, n_0} C_{n, n_0}^\top V_\perp ) 
}{  \bv^\top C_{n, n_0} C_{n, n_0}^\top \bv},$$
where $c$ is an absolute constant.
\label{lemma: change}
\end{lemma}

Define the step sizes $\hat{\eta}_t = \eta_{t+n_0} = \frac{1}{n_0 + t}$ for all positive integer $t$.  We can see that
\begin{align*}
C_{n, n_0} &= (I +\eta_{n_0+n} A_{n_0+n}) \cdots (I +\eta_{n_0+1} A_{n_0+1})\\
 &= (I +\hat{\eta}_n A_{n_0+n}) \cdots (I +\hat{\eta}_1 A_{n_0+1}).
\end{align*}
Applying $C_{n, n_0}$ on $\bw$ gives us $$\sin^2(\bv, \frac{C_{n, n_0}\bw}{\|C_{n, n_0}\bw\|_2})  \leq \frac{c_1 \log (1 / \delta)}{\delta} \frac{  Tr(V_\perp^\top C_{n, n_0} C_{n, n_0}^\top V_\perp ) 
}{  \bv^\top C_{n, n_0} C_{n, n_0}^\top \bv  },$$ where $c_1$ is some absolute constant.
Therefore, Lemma \ref{lemma: change} can be interpreted as a relationship between the error bound after $n+n_0$ iterations of Oja's algorithm and the error bound of applying the last $n$ iterations on a randomly chosen initial vector if we totally dump the result from the first $n_0$ iterations. With Lemma \ref{lemma: change} and a carefully selected constant integer $n_0$, we can prove Theorem~\ref{thm: our}, which implies Theorem~\ref{thm: main} directly. 

\begin{theorem}
Fix any $\delta > 0$. Let $n_0$ be the smallest integer such that $n_0 > \max( 4 \mathcal{M}, 18 \frac{\mathcal{V} +\lambda_1^2}{\log (1 + \frac{\delta}{100})}).$ Assume $B_{n+n_0}$ is invertible, the output $\bw_{n +n_0}$ of the History PCA algorithm (rank-$1$ with $m=1$) satisfies:
\begin{align*}
1 - (\bw_{n +n_0}^\top  v_1)^2 &= O( \frac{\mathcal{V}}{|2(\lambda_1 - \lambda_2) - 1|}\frac{1}{n +n_0}),
\end{align*}
with probability at least $1 - \delta$. 
\label{thm: our}
\end{theorem}


\section{Experimental Results}
\label{sec:exp}
In this section we conduct numerical experiments to compare our algorithm with previous 
streaming PCA approaches using synthetic streaming data as well as real world data listed in Table~\ref{tab:datasets}.

\subsection{Choice of $m$ for History PCA}
In Figure~\ref{fig:m}, we compare the performances of our History PCA algorithm with different choices of $m$ (number of inner iterations) on NIPS data set with $k = 1$. Here the accuracy is the explained variance of the algorithm output. We can see that when $m = 3, 5, 7$, History PCA's performances are equally well. And when $m = 1$, its performance is slightly worse. Thus, a small $m$ is already good enough for our algorithm. 
In the following experiments, we will just set  $m=3$ in our algorithm to compare with other existing streaming PCA methods.

\begin{figure}[]
\vspace*{-0.2in}
  \centering
    \includegraphics[width=0.6\linewidth]{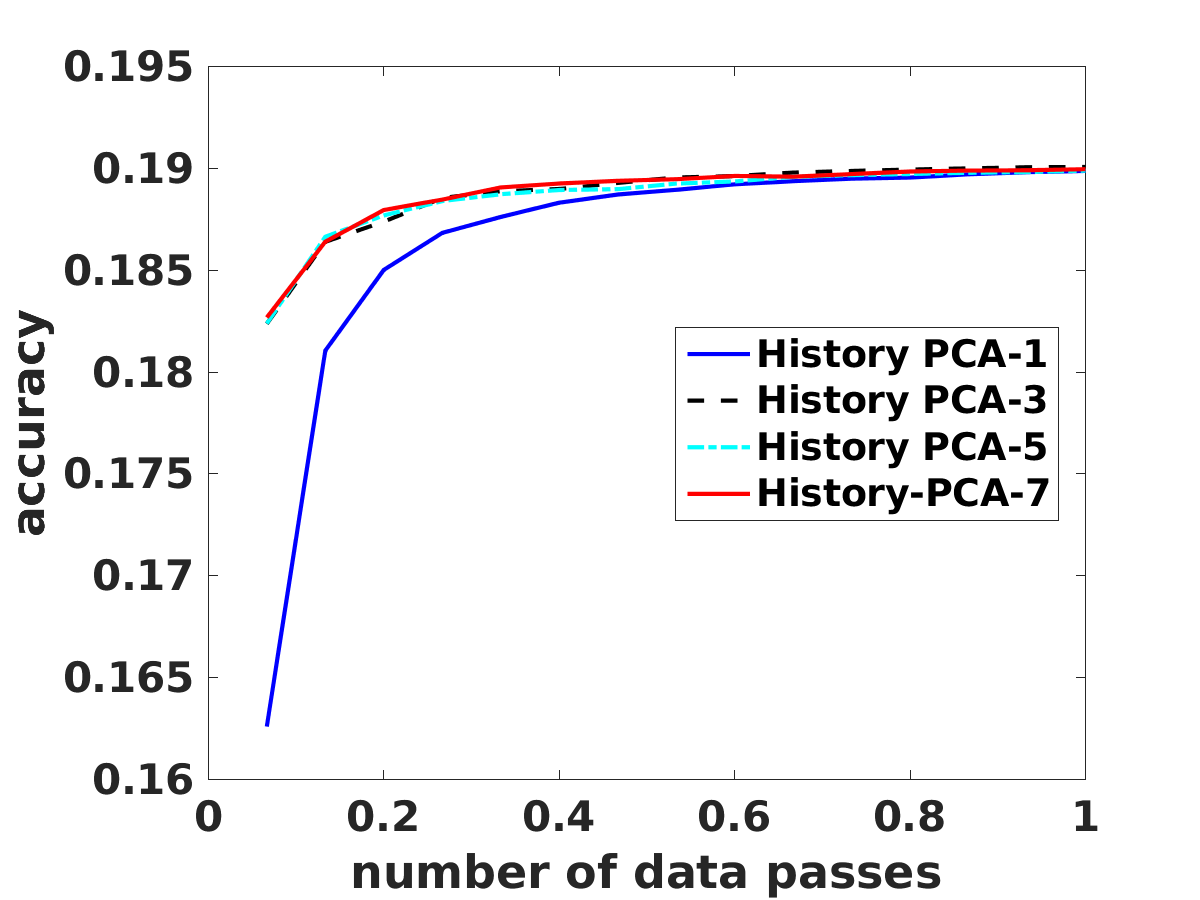}
    \label{fig:m}
  \caption{The choice of $m$ in History PCA on NIPS data set} 
  \label{fig:m}
\end{figure}

\subsection{Simulated Streaming Data}

In this section, we compare the proposed History PCA algorithm with state-of-the-art streaming 
PCA algorithms: block power method, DBPCA method, Oja's algorithm and Oja$^{++}$ algorithm. 

\begin{figure}[H]
\vspace*{-0.1in}
  \centering 
  \resizebox{0.98\linewidth}{!}{
  \begin{tabular}{ccc}
        \subfloat[$d = 100, B = 10, k = 1, sig = 0.1$ \label{fig:Oja++n10000d100B10k1iter1sigma0.1}]{
    \includegraphics[width=0.32\linewidth]{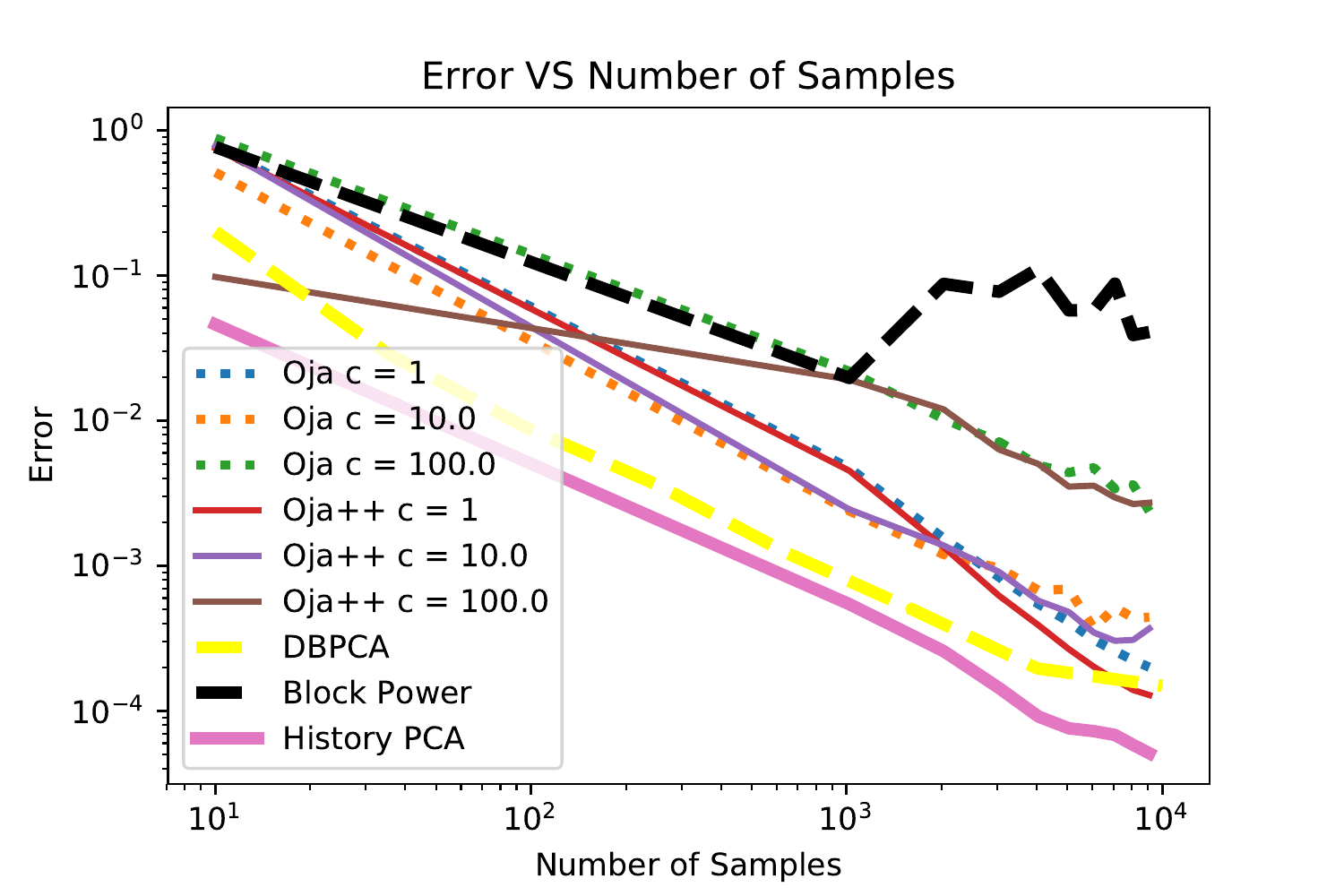}
    \label{fig:compare}
    }& \hspace{-10pt}
    \subfloat[$d = 100, B = 10, k = 1, sig = 0.5$ \label{fig:n20000_d1000_B100_k5_iter1}]{
    \includegraphics[width=0.32\linewidth]{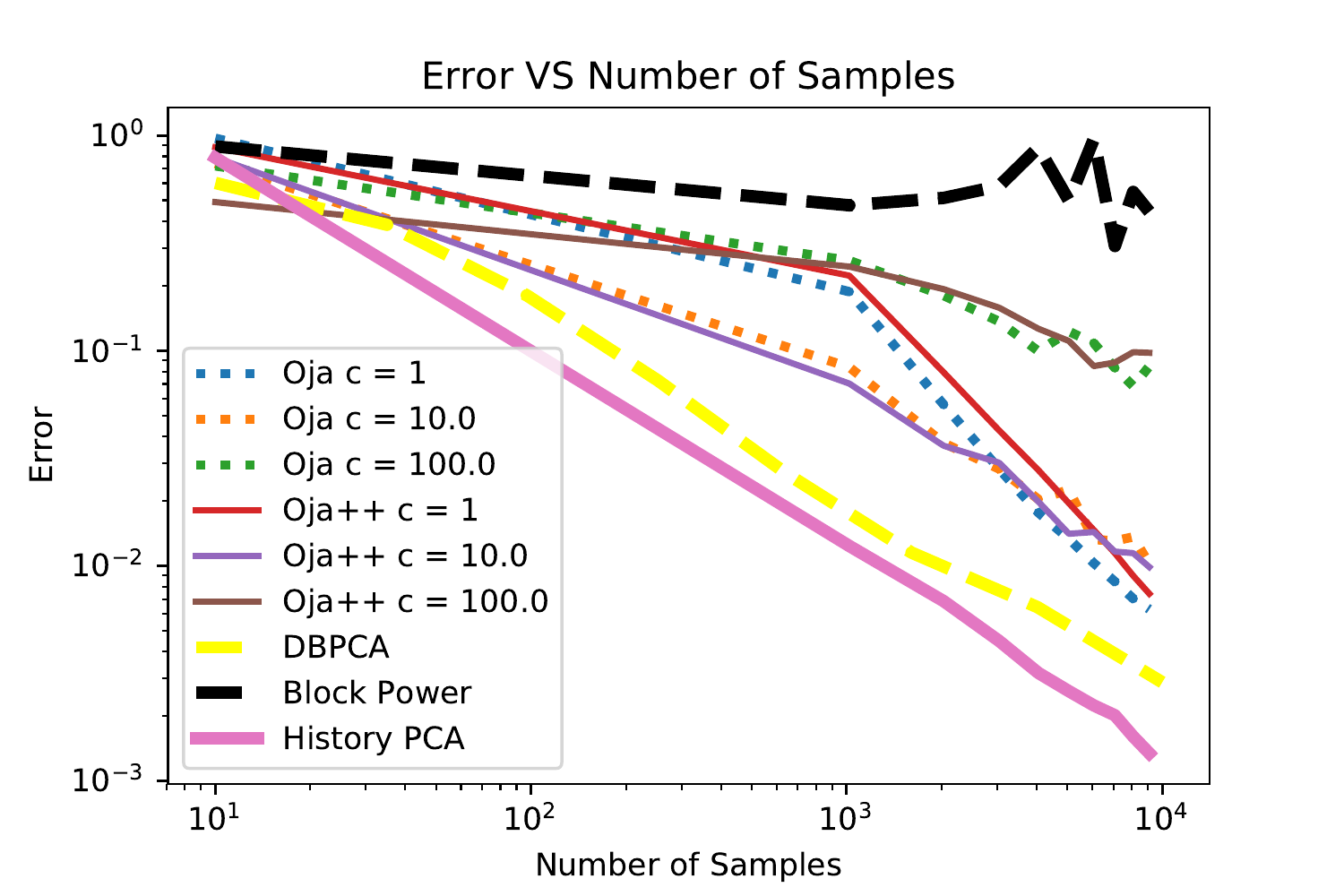}
    \label{fig:timevspre}
    }& \hspace{-10pt}
    \subfloat[$d = 100, B = 10, k = 1, sig = 0.8$ \label{fig:n10000_d100_B100_k10_iter1}]{
    \includegraphics[width=0.32\linewidth]{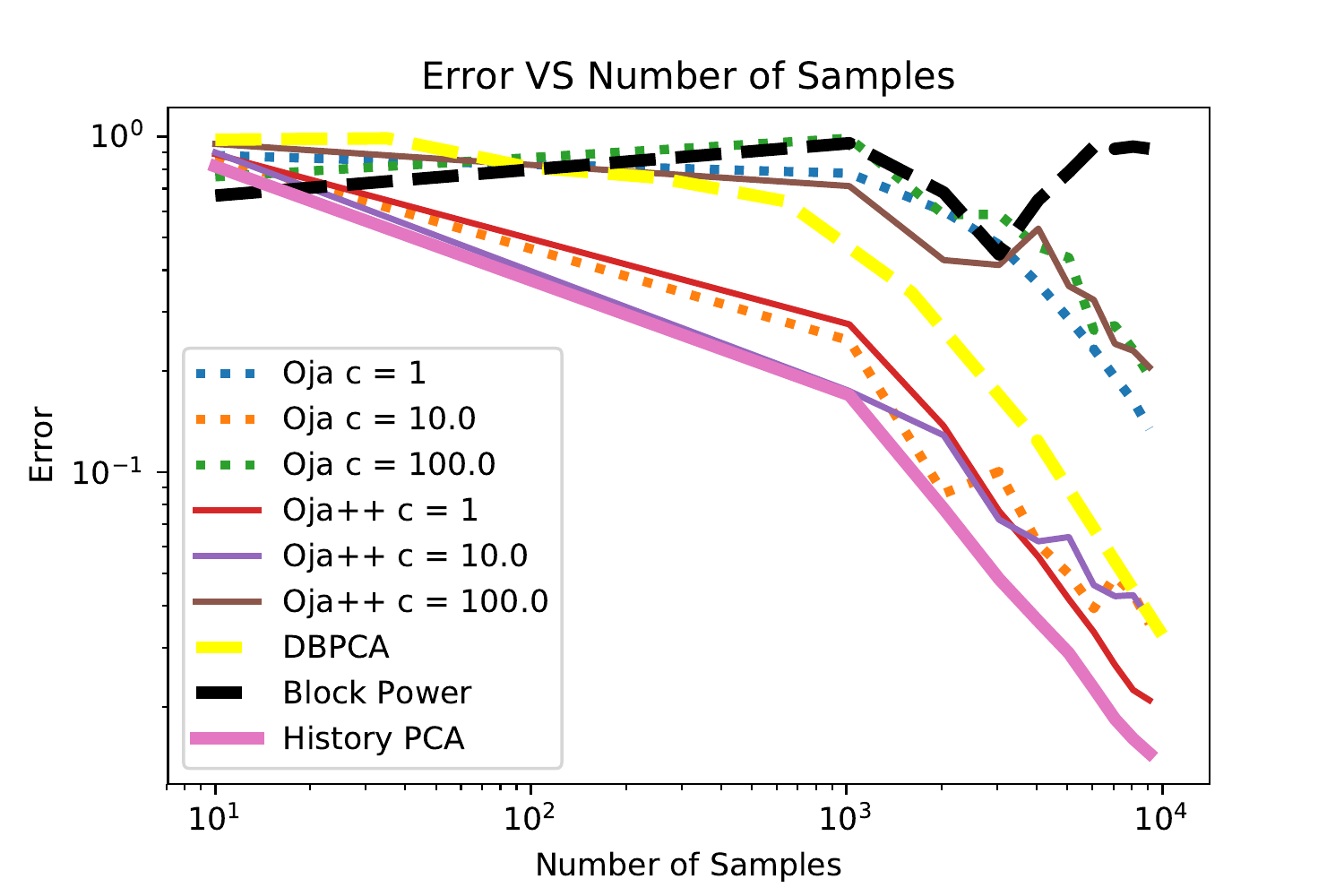}
    \label{fig:timevspre}
    }\\
   \hspace{-10pt}
    \subfloat[$d = 100, B = 10, k = 5, sig = 0.1$ \label{fig:Oja++n10000d100B10k1iter1sigma0.1}]{
    \includegraphics[width=0.32\linewidth]{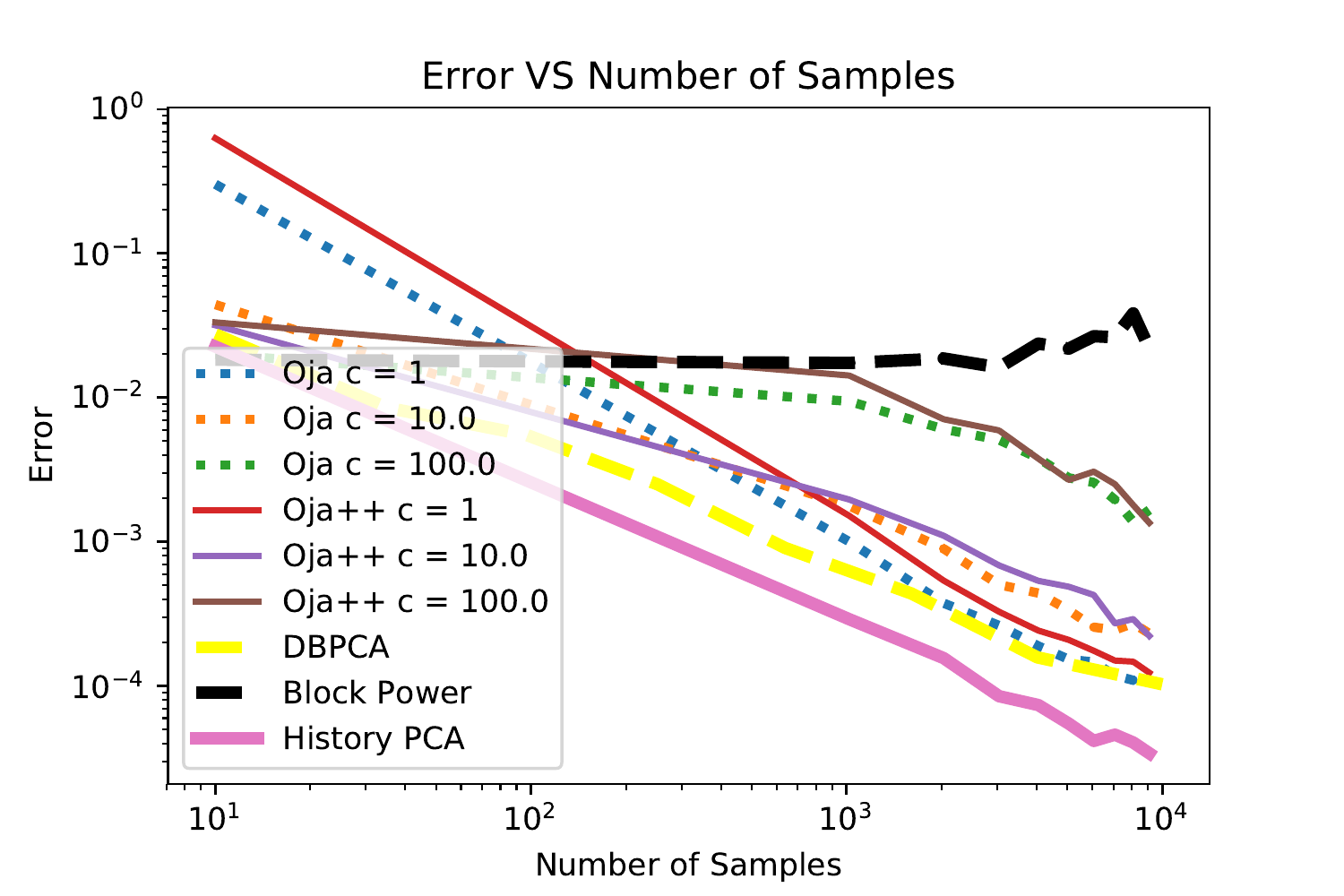}
    \label{fig:compare}
    }& \hspace{-10pt}
    \subfloat[$d = 100, B = 10, k = 5, sig = 0.5$ \label{fig:n20000_d1000_B100_k5_iter1}]{
    \includegraphics[width=0.32\linewidth]{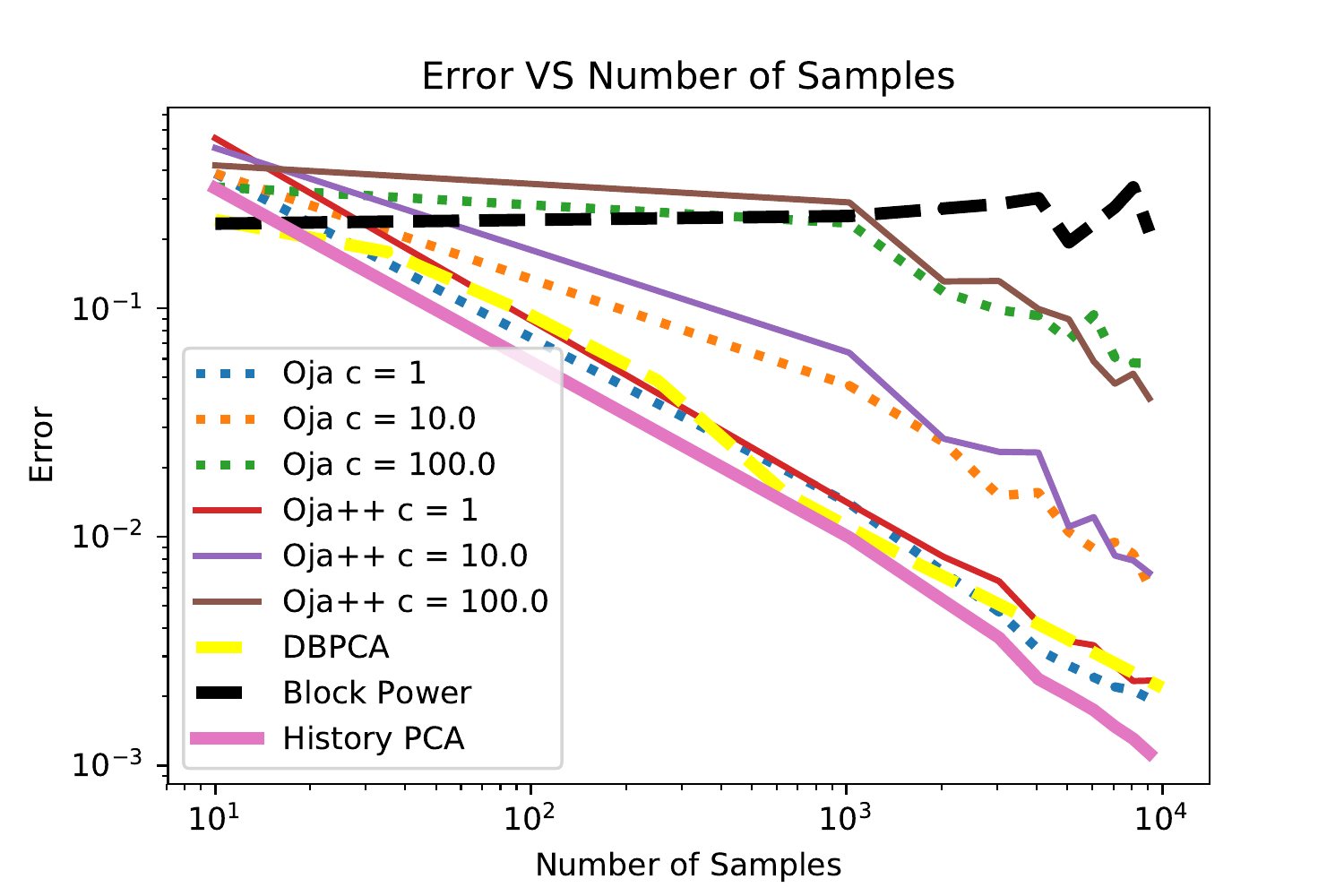}
    \label{fig:timevspre}
    }& \hspace{-10pt}
    \subfloat[$d = 100, B = 10, k = 5, sig = 0.8$ \label{fig:n10000_d100_B100_k10_iter1}]{
    \includegraphics[width=0.32\linewidth]{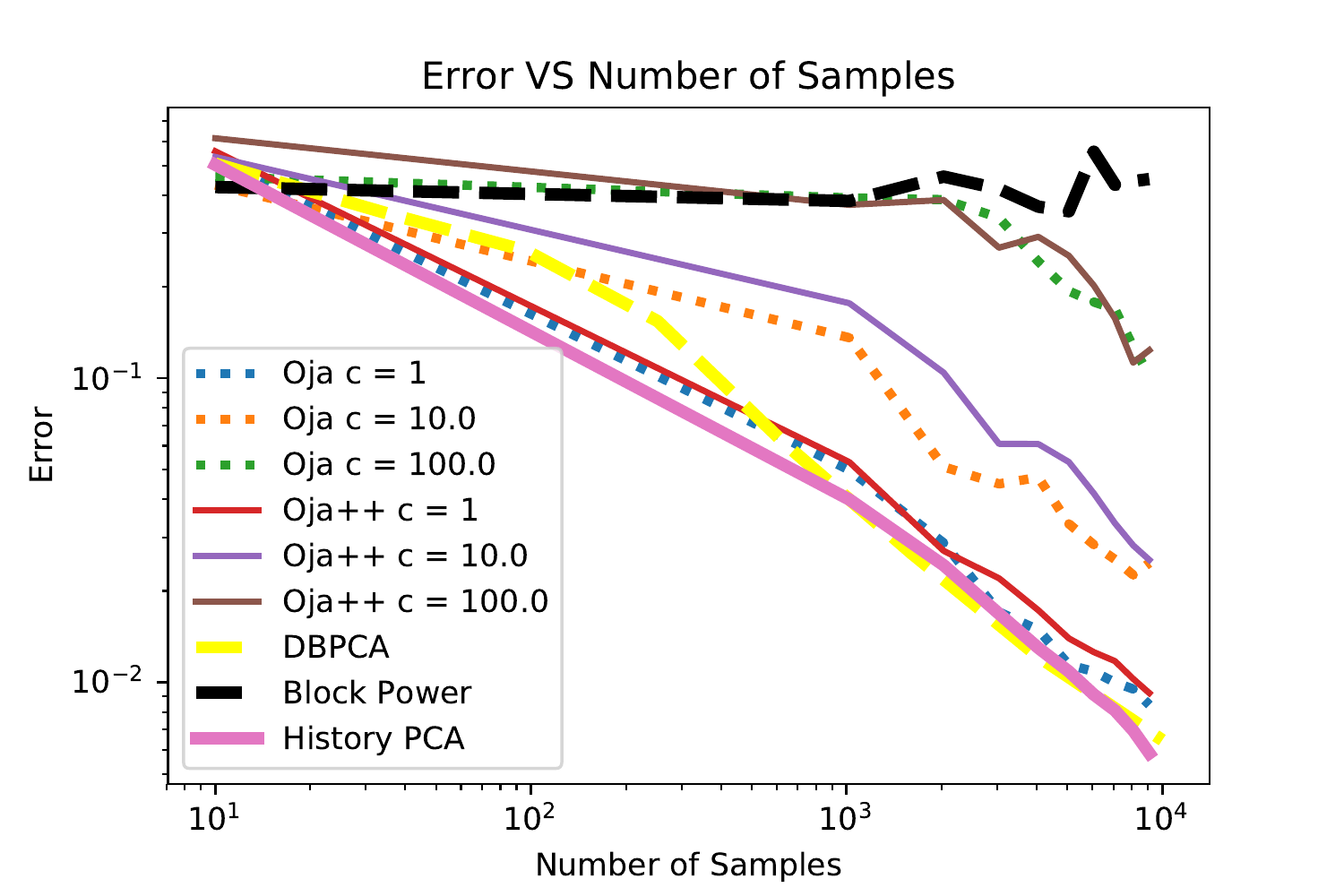}
    \label{fig:timevspre}
    }\\
   \hspace{-10pt}
    \subfloat[$d = 100, B = 100, k = 1, sig = 0.1$ \label{fig:Oja++n10000d100B100k1iter1sigma0.1}]{
    \includegraphics[width=0.32\linewidth]{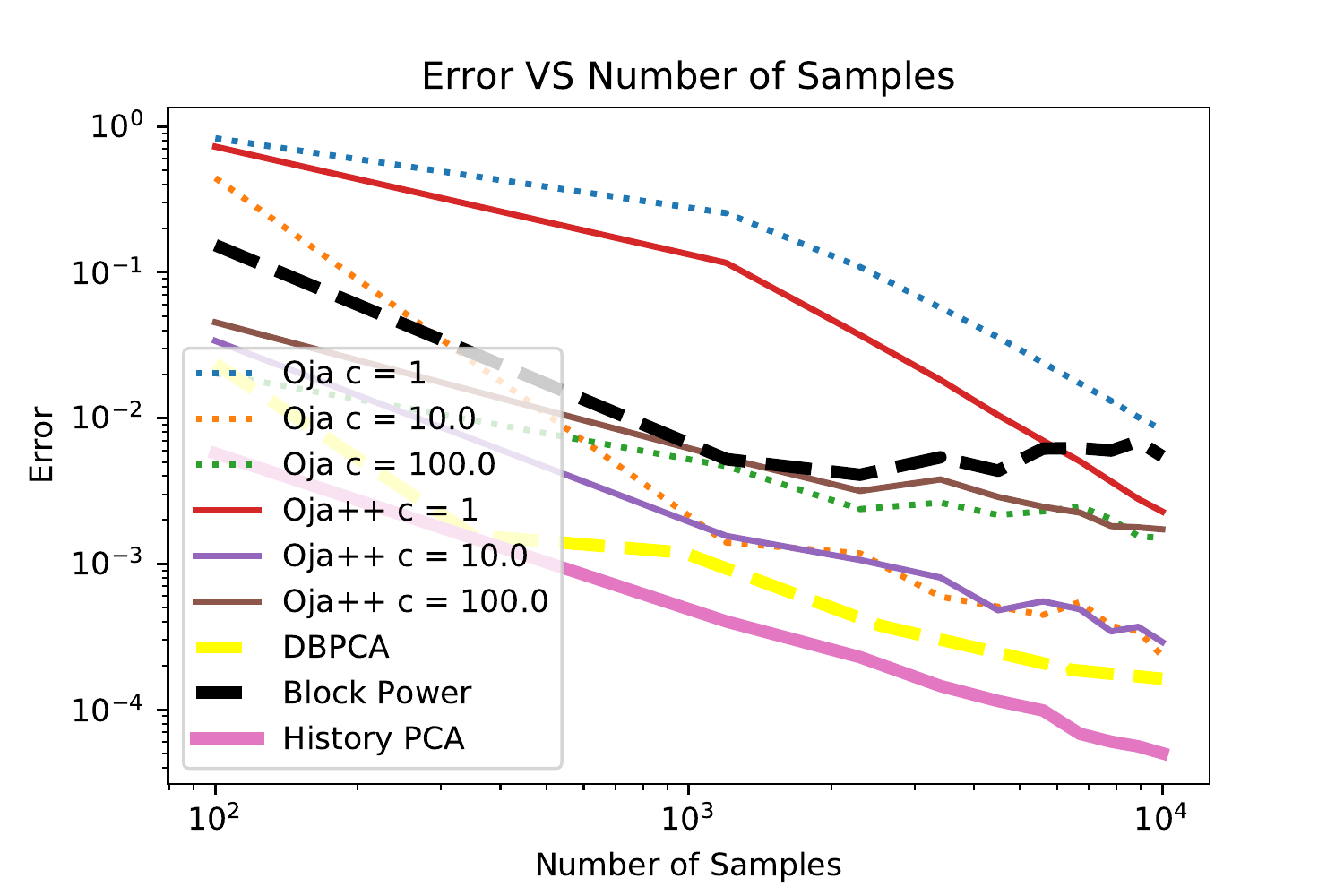}
    \label{fig:compare}
    }& \hspace{-10pt}
    \subfloat[$d = 100, B = 100, k = 1, sig = 0.5$ \label{fig:n10000_d100_B100_k5_iter1}]{
    \includegraphics[width=0.32\linewidth]{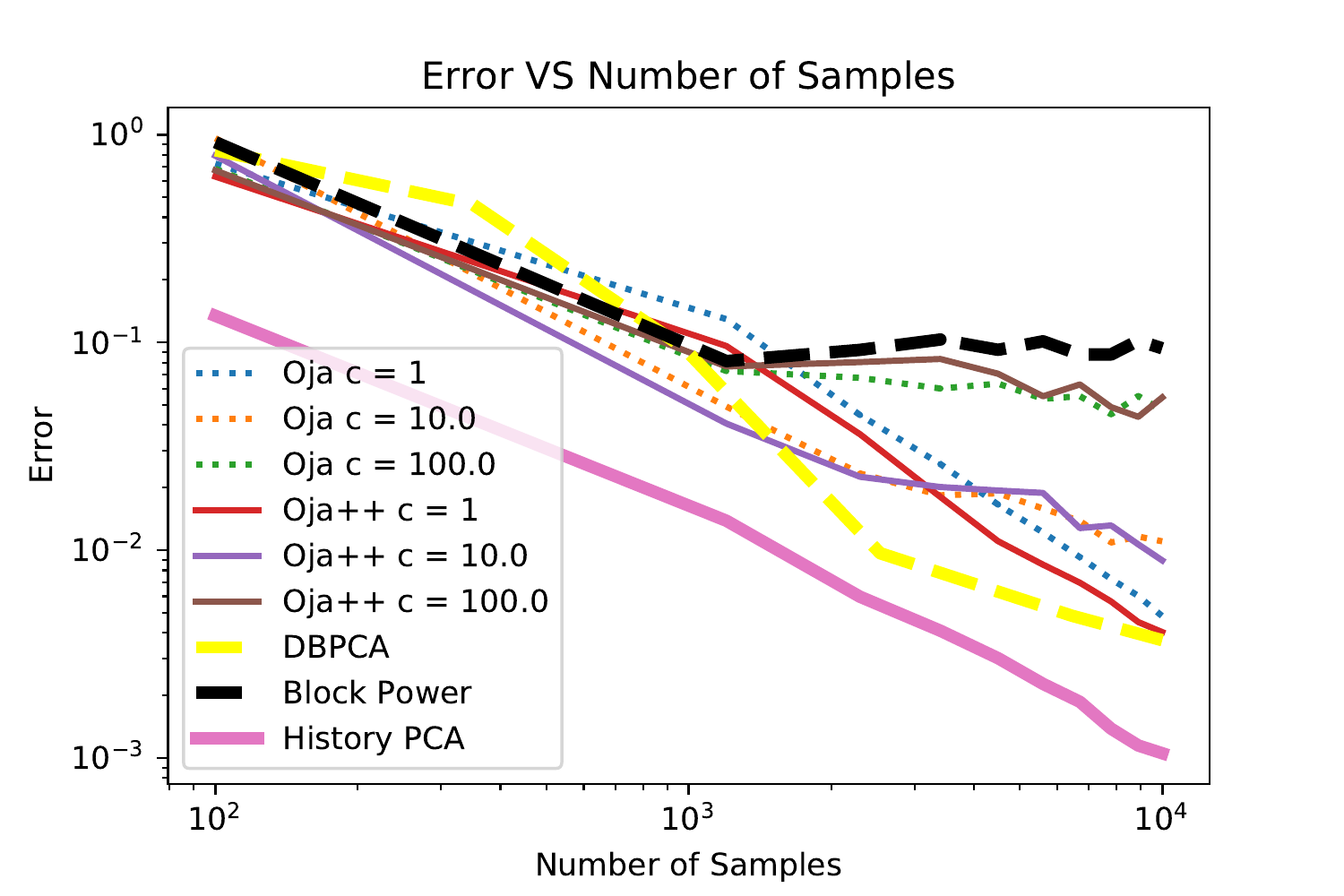}
    \label{fig:timevspre}
    }& \hspace{-10pt}
    \subfloat[$d = 100, B = 100, k = 1, sig = 0.8$ \label{fig:n10000_d100_B100_k10_iter1}]{
    \includegraphics[width=0.32\linewidth]{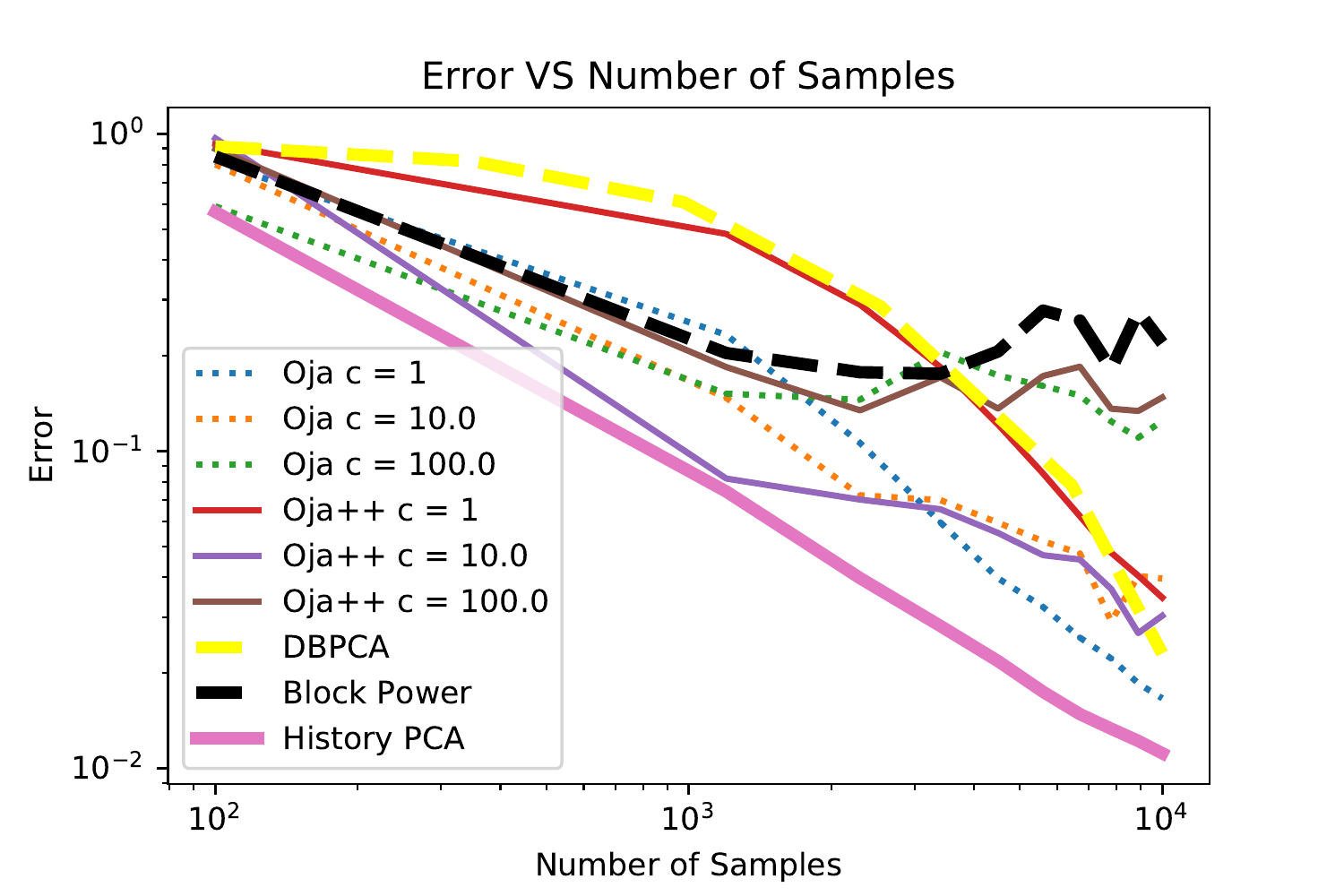}
    \label{fig:timevspre}
    }\\
   \hspace{-10pt}
    \subfloat[$d = 100, B = 100, k = 5, sig = 0.1$ \label{fig:n10000_d1000_B100_k1_iter1}]{
    \includegraphics[width=0.32\linewidth]{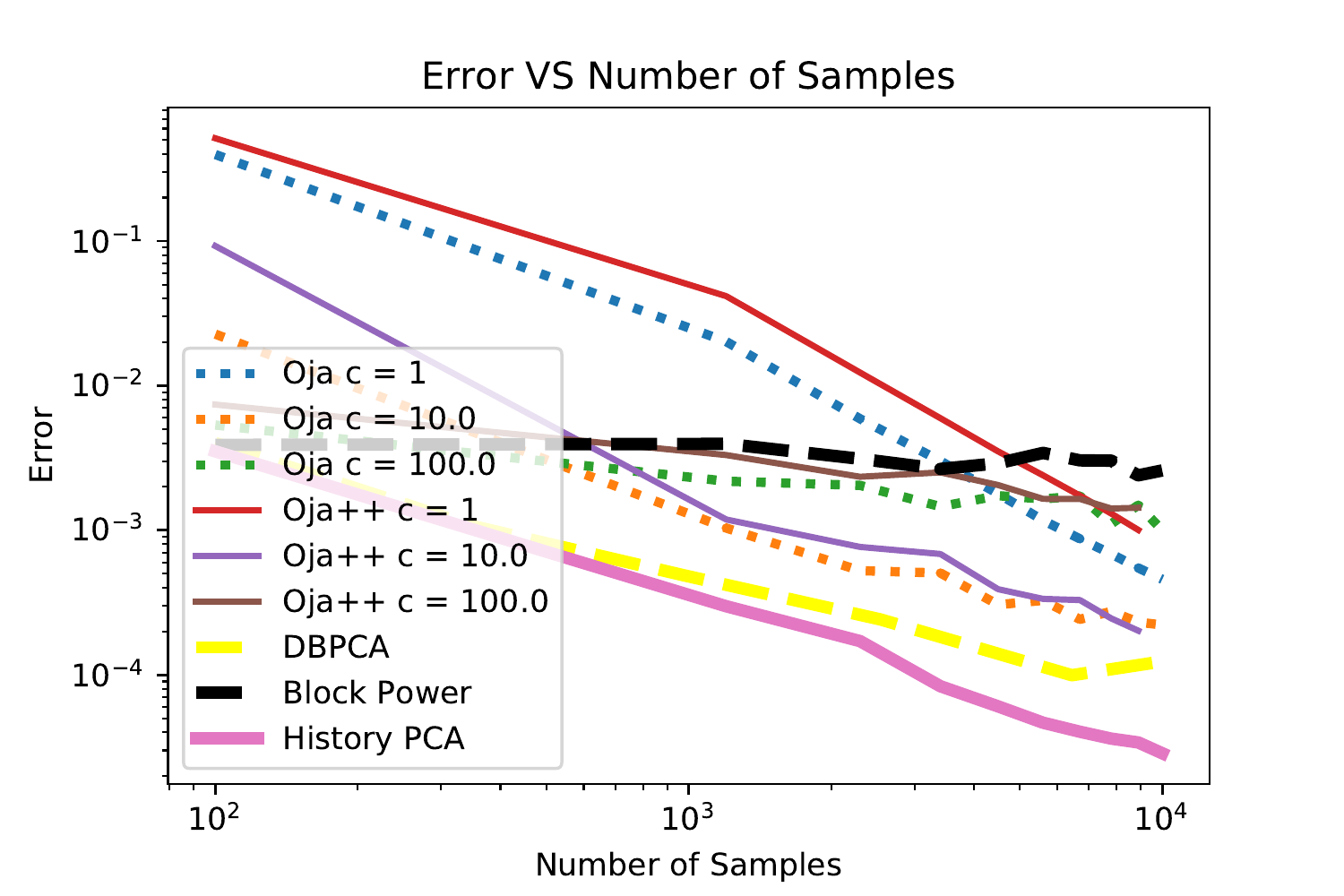}
    \label{fig:meovspre}
    } & \hspace{-10pt}
    \subfloat[$d = 100, B = 100, k = 5, sig = 0.5$ \label{fig:n10000_d1000_B100_k5_iter1}]{
    \includegraphics[width=0.32\linewidth]{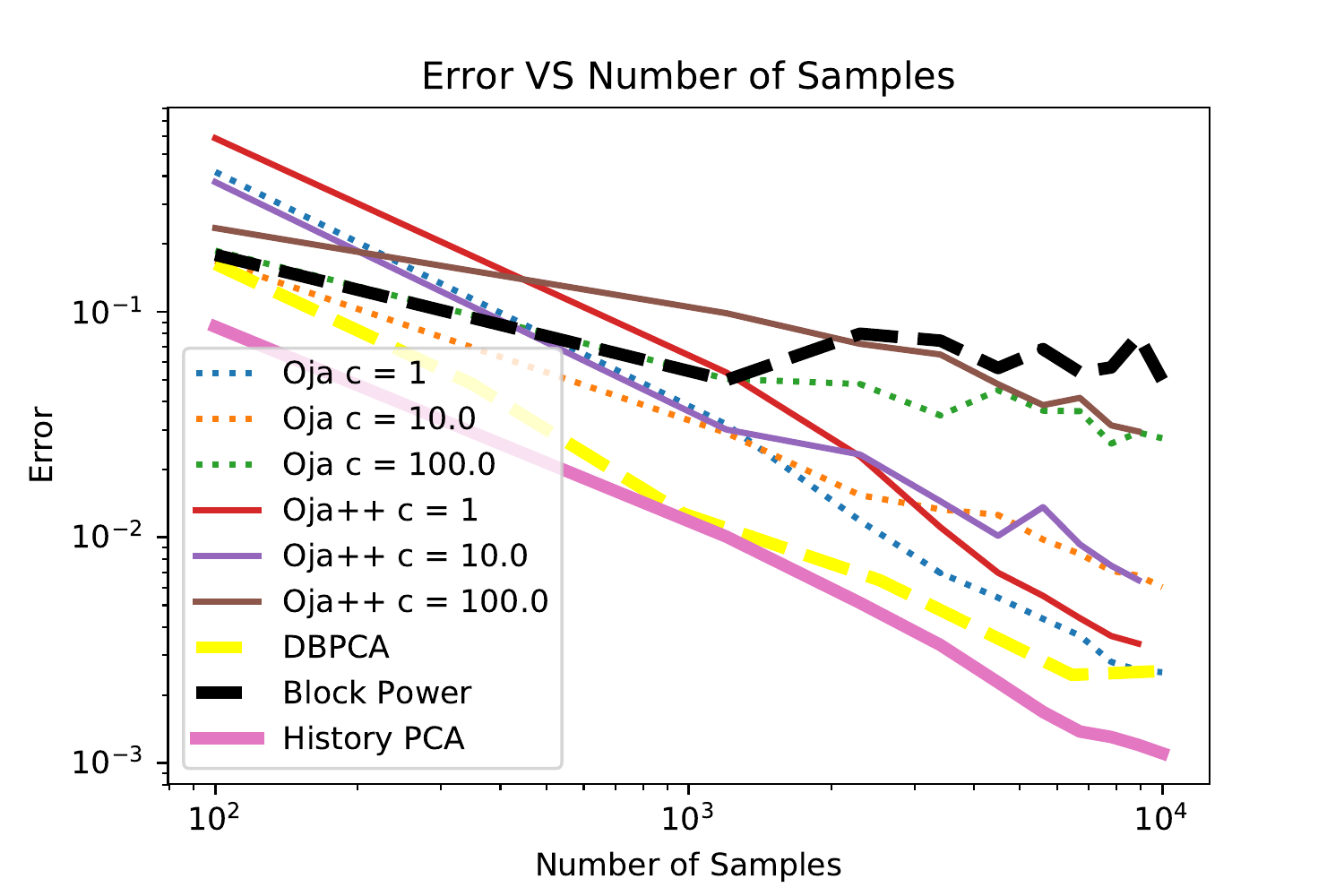}
    \label{fig:meovspre}
    }& \hspace{-10pt}
    \subfloat[$d = 100, B = 100, k = 5, sig = 0.8$\label{fig:n10000_d1000_B100_k10_iter1}]{
    \includegraphics[width=0.32\linewidth]{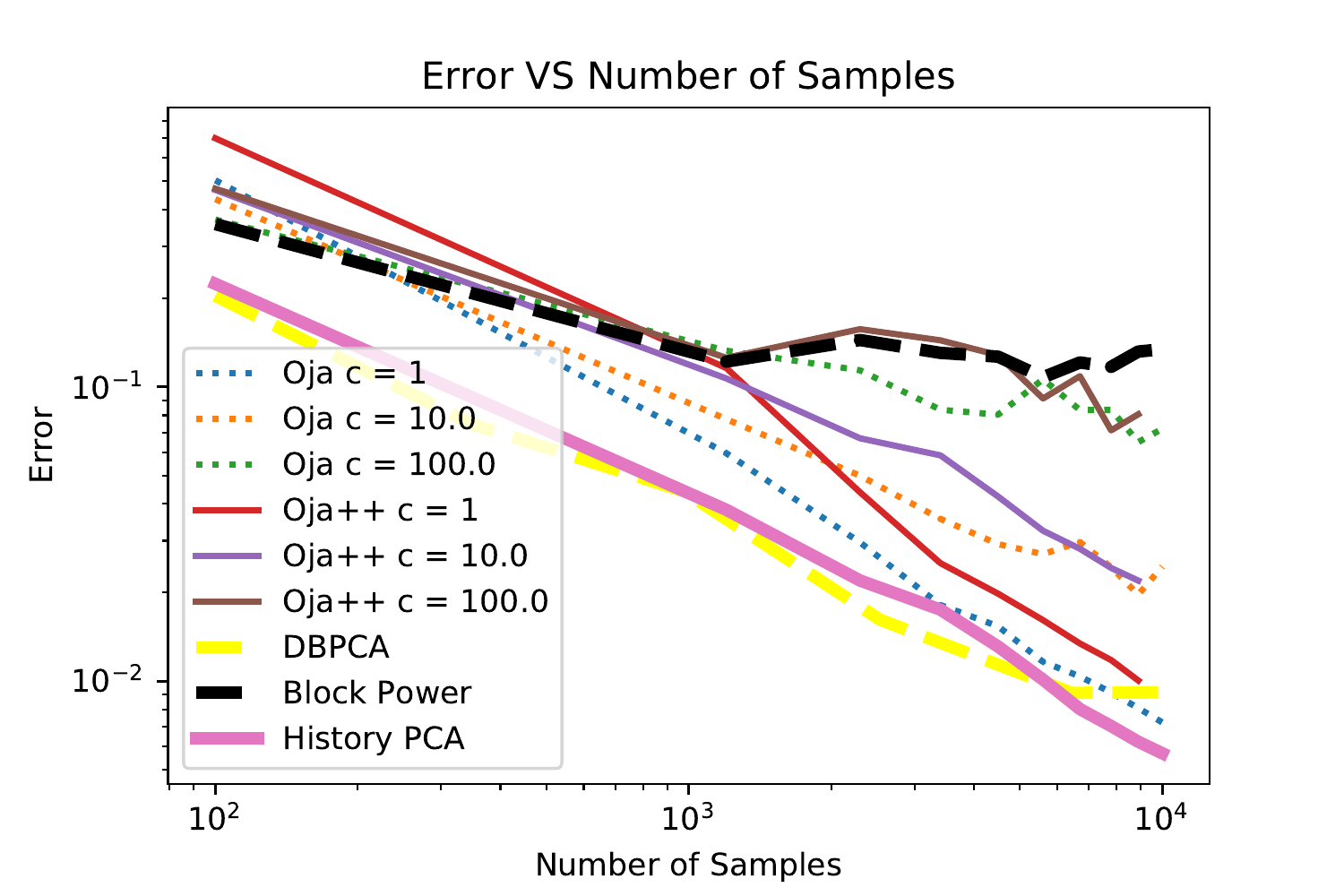}    \label{fig:timevspre}
    }\\
       \hspace{-10pt}
    \subfloat[$d = 100, B = 100, k = 10, sig = 0.1$ \label{fig:n10000_d1000_B100_k1_iter1}]{
    \includegraphics[width=0.32\linewidth]{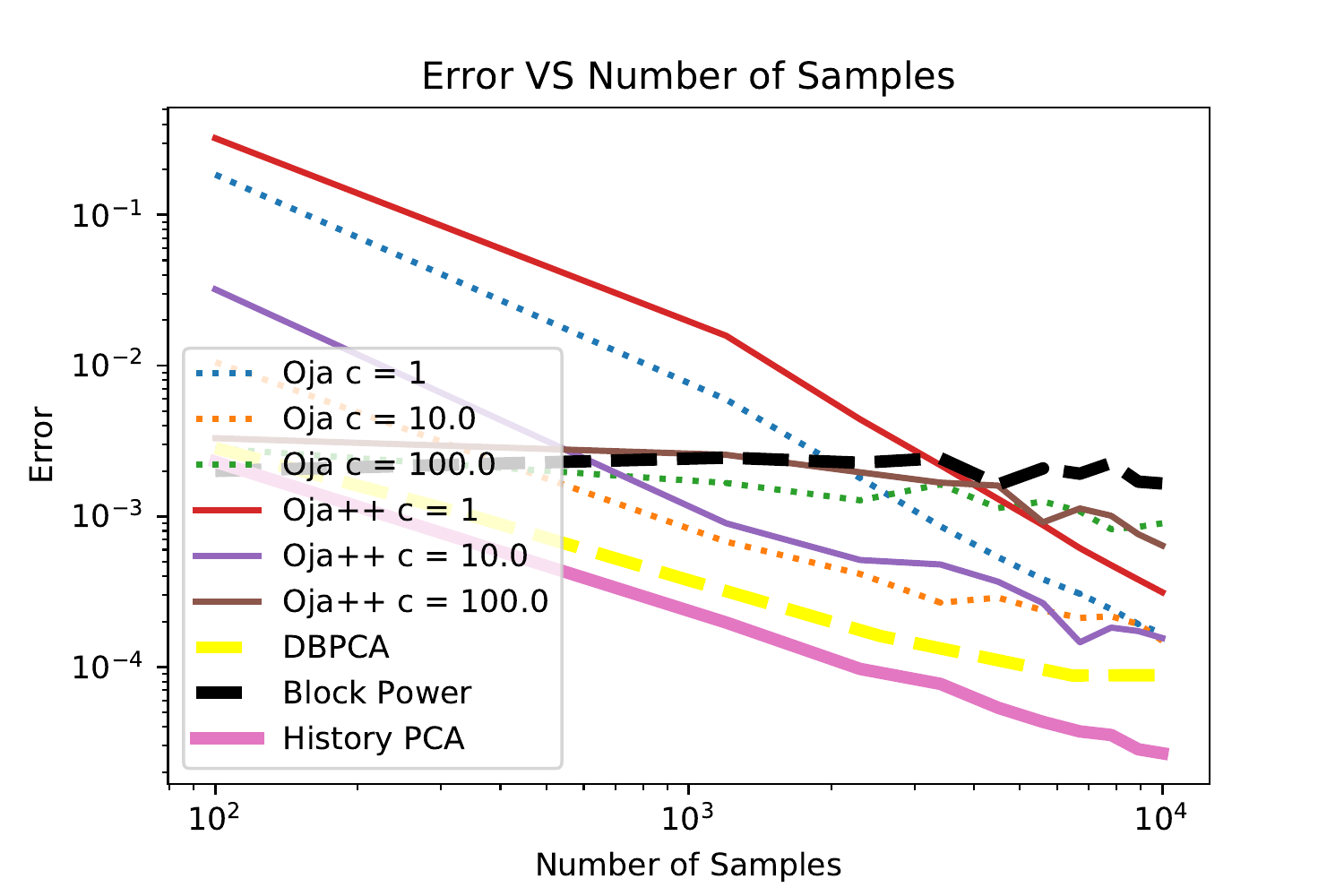}
    \label{fig:meovspre}
    } & \hspace{-10pt}
    \subfloat[$d = 100, B = 100, k = 10, sig = 0.5$ \label{fig:n10000_d1000_B100_k5_iter1}]{
    \includegraphics[width=0.32\linewidth]{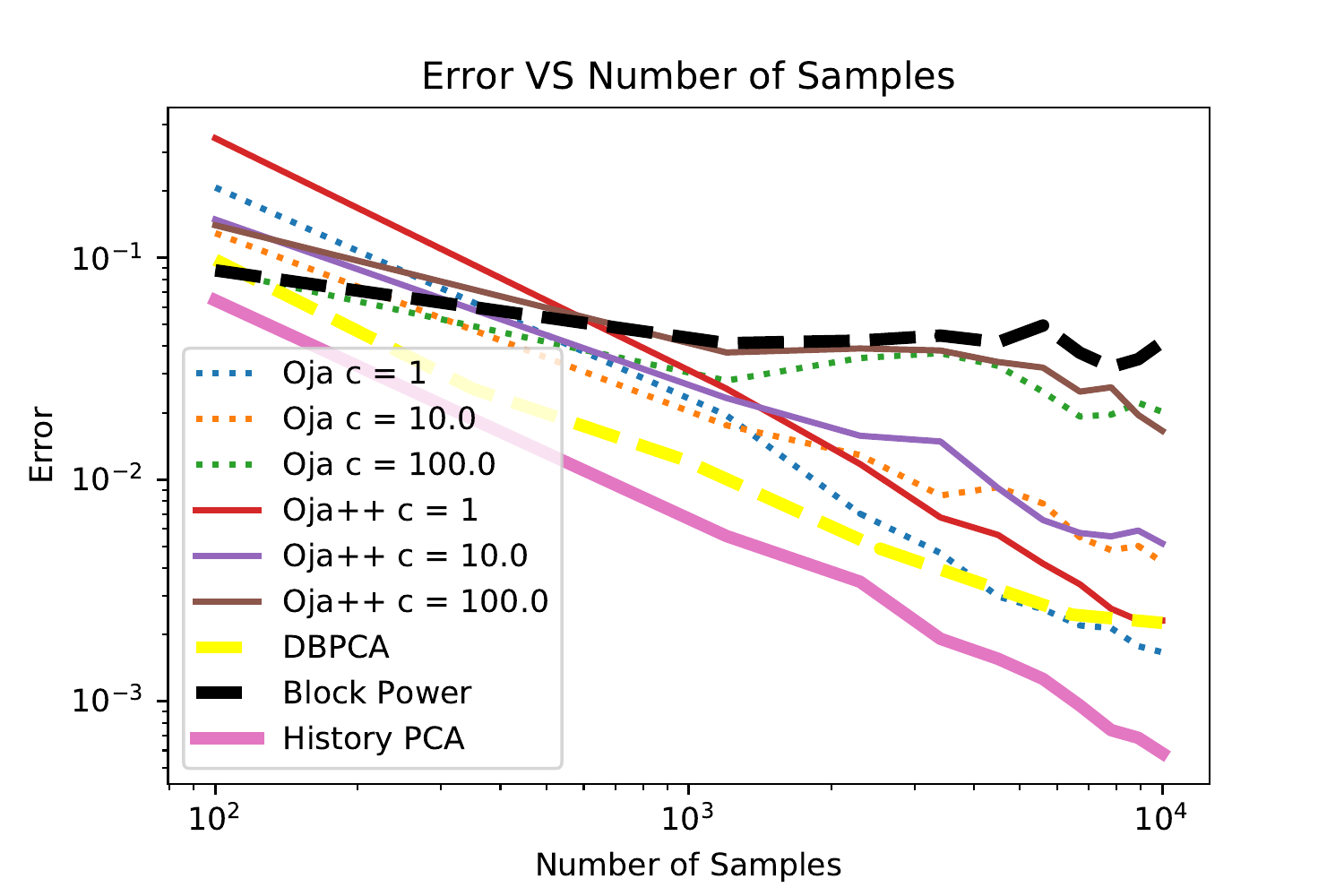}
    \label{fig:meovspre}
    }& \hspace{-10pt}
    \subfloat[$d = 100, B = 100, k = 10, sig = 0.8$\label{fig:n10000_d1000_B100_k10_iter1}]{
    \includegraphics[width=0.32\linewidth]{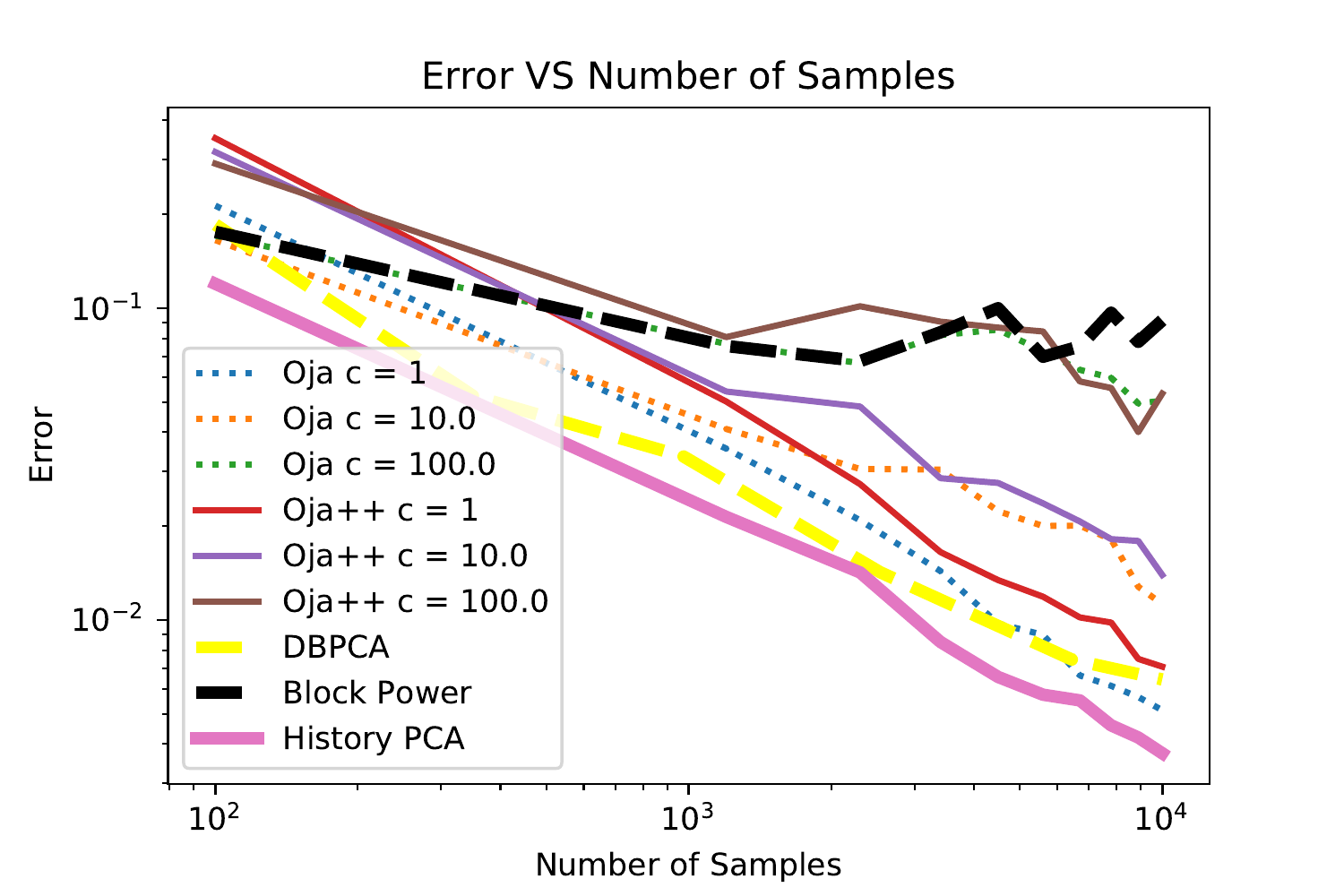}    \label{fig:timevspre}
    }
  \end{tabular} 
  }
  \caption{Comparison of streaming PCA algorithms on simulated data sets (d = 100).} 
  \label{fig: simulated_data2}
\end{figure}

\begin{figure*}[t]
\vspace*{-0.1in}
  \centering
  \begin{tabular}{ccc}
    \subfloat[$d = 1000, B = 100, k = 1, sig = 0.1$ \label{fig:Oja++n10000d100B100k1iter1sigma0.1}]{
    \includegraphics[width=0.32\linewidth]{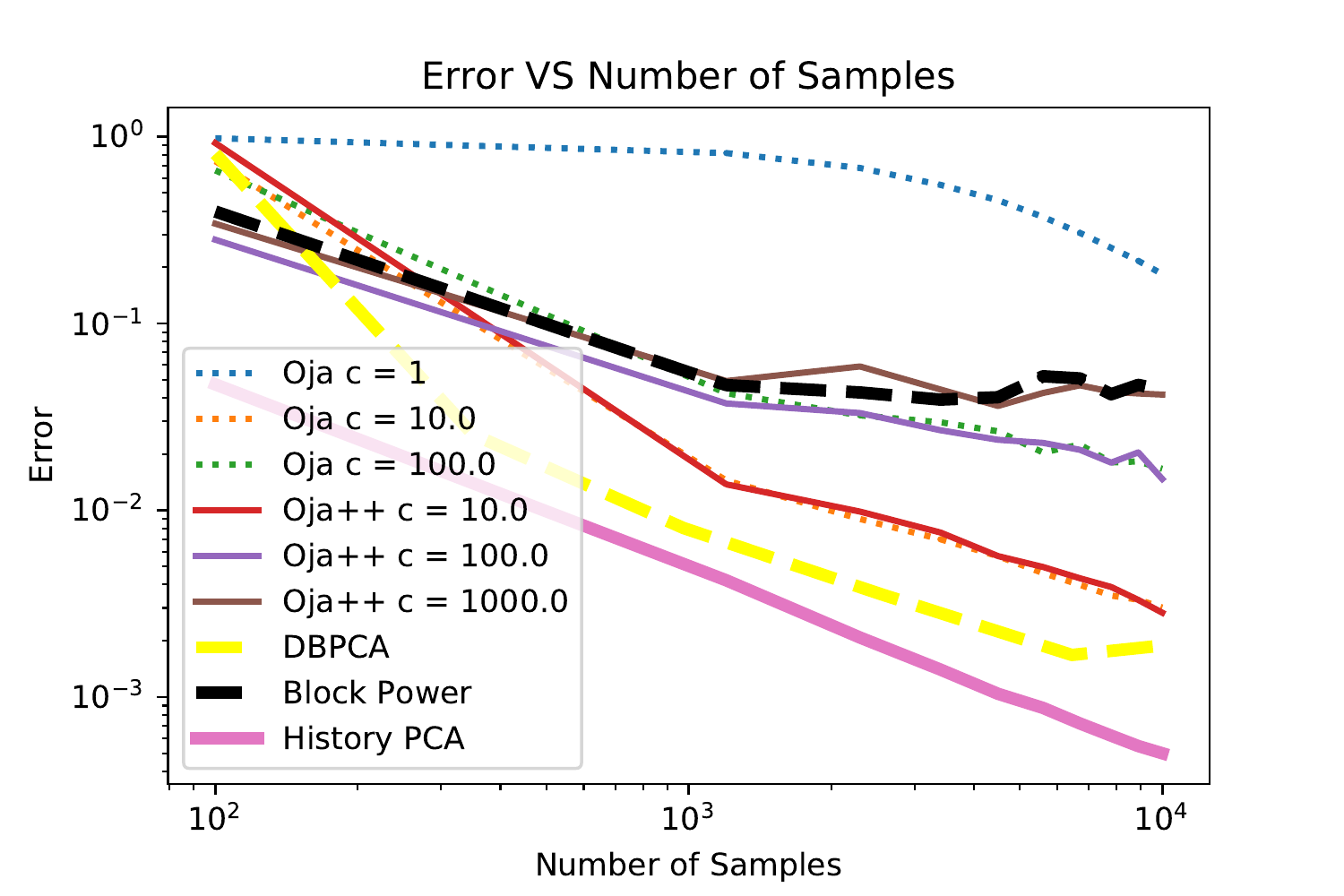}
    \label{fig:compare}
    }& \hspace{-10pt}
    \subfloat[$d = 1000, B = 100, k = 1, sig = 0.5$ \label{fig:n20000_d1000_B1000_k5_iter1}]{
    \includegraphics[width=0.32\linewidth]{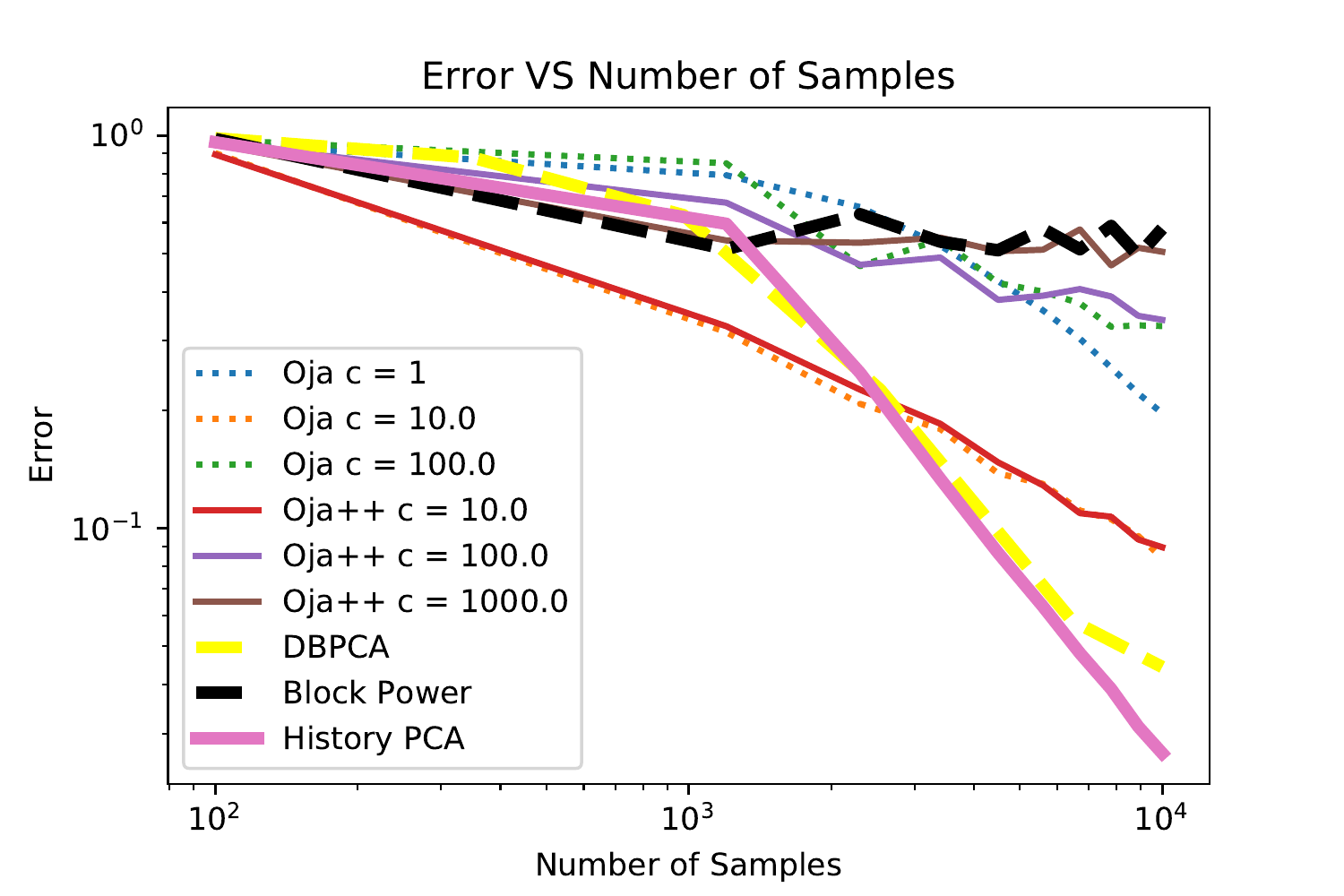}
    \label{fig:timevspre}
    }& \hspace{-10pt}
    \subfloat[$d = 1000, B = 100, k = 1, sig = 0.8$ \label{fig:n10000_d100_B100_k10_iter1}]{
    \includegraphics[width=0.32\linewidth]{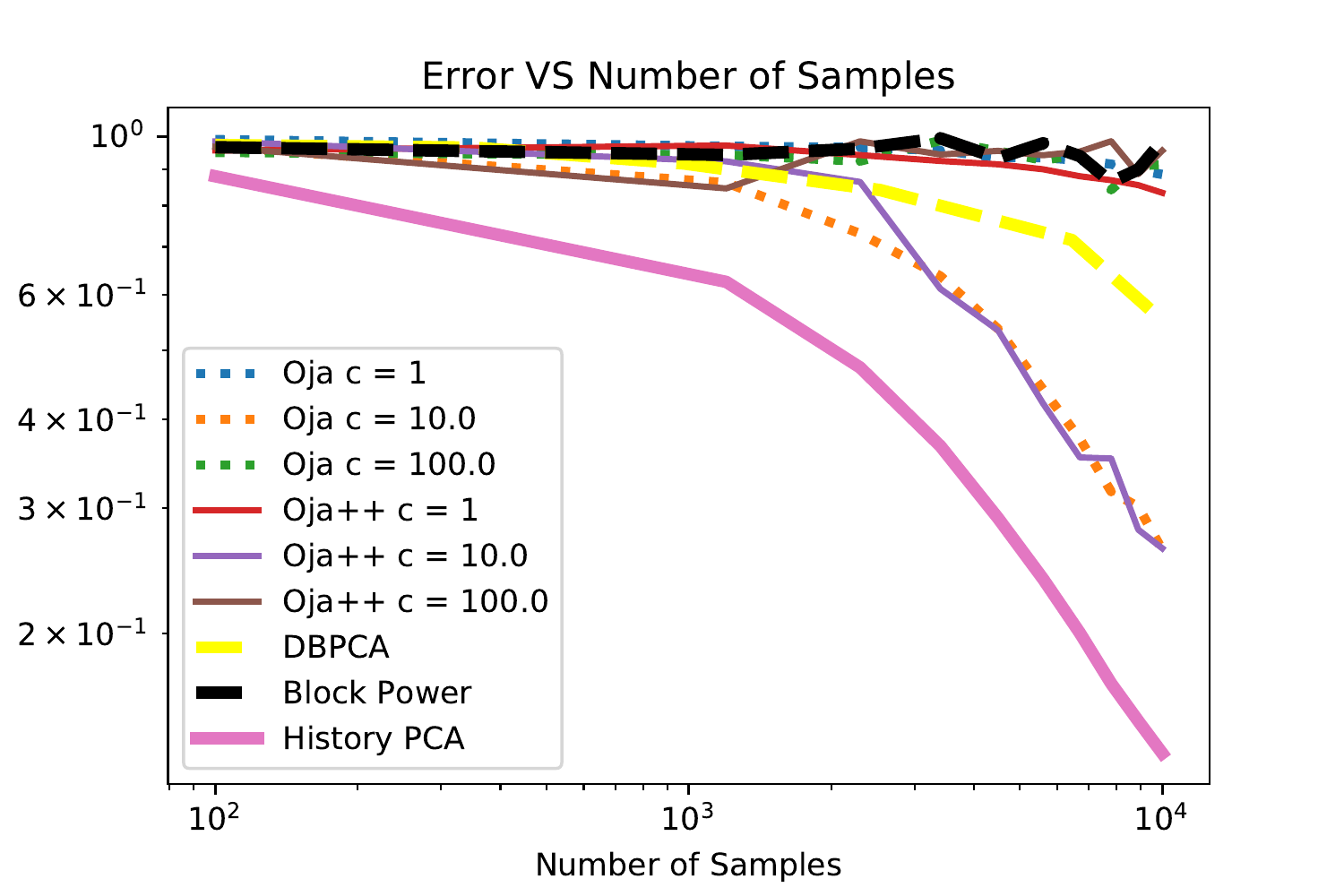}
    \label{fig:timevspre}
    }\\
   \hspace{-10pt}
    \subfloat[$d = 1000, B = 100, k = 5, sig = 0.1$ \label{fig:n10000_d1000_B100_k1_iter1}]{
    \includegraphics[width=0.32\linewidth]{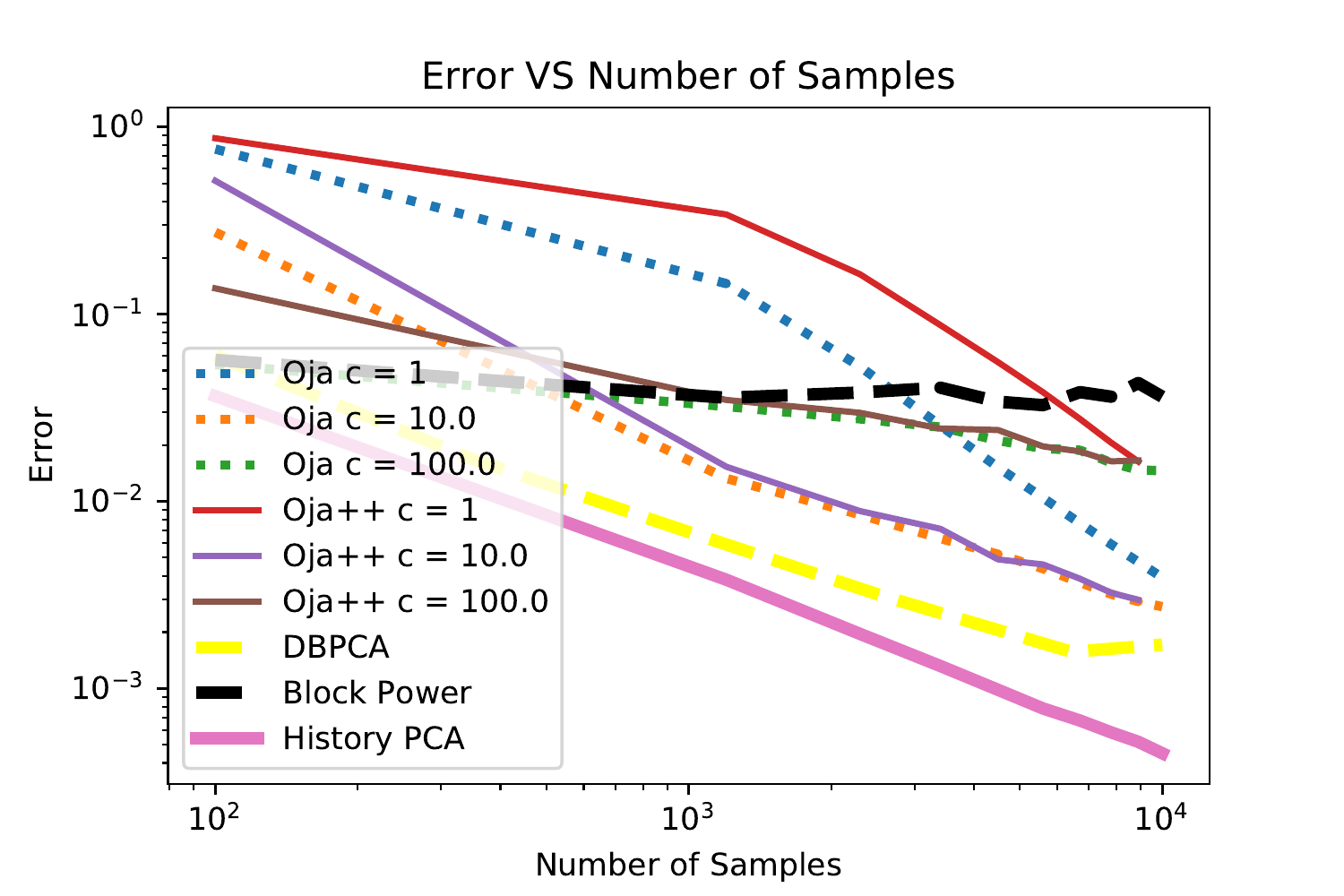}
    \label{fig:meovspre}
    } & \hspace{-10pt}
    \subfloat[$d = 1000, B = 100, k = 5, sig = 0.5$ \label{fig:n10000_d1000_B100_k5_iter1}]{
    \includegraphics[width=0.32\linewidth]{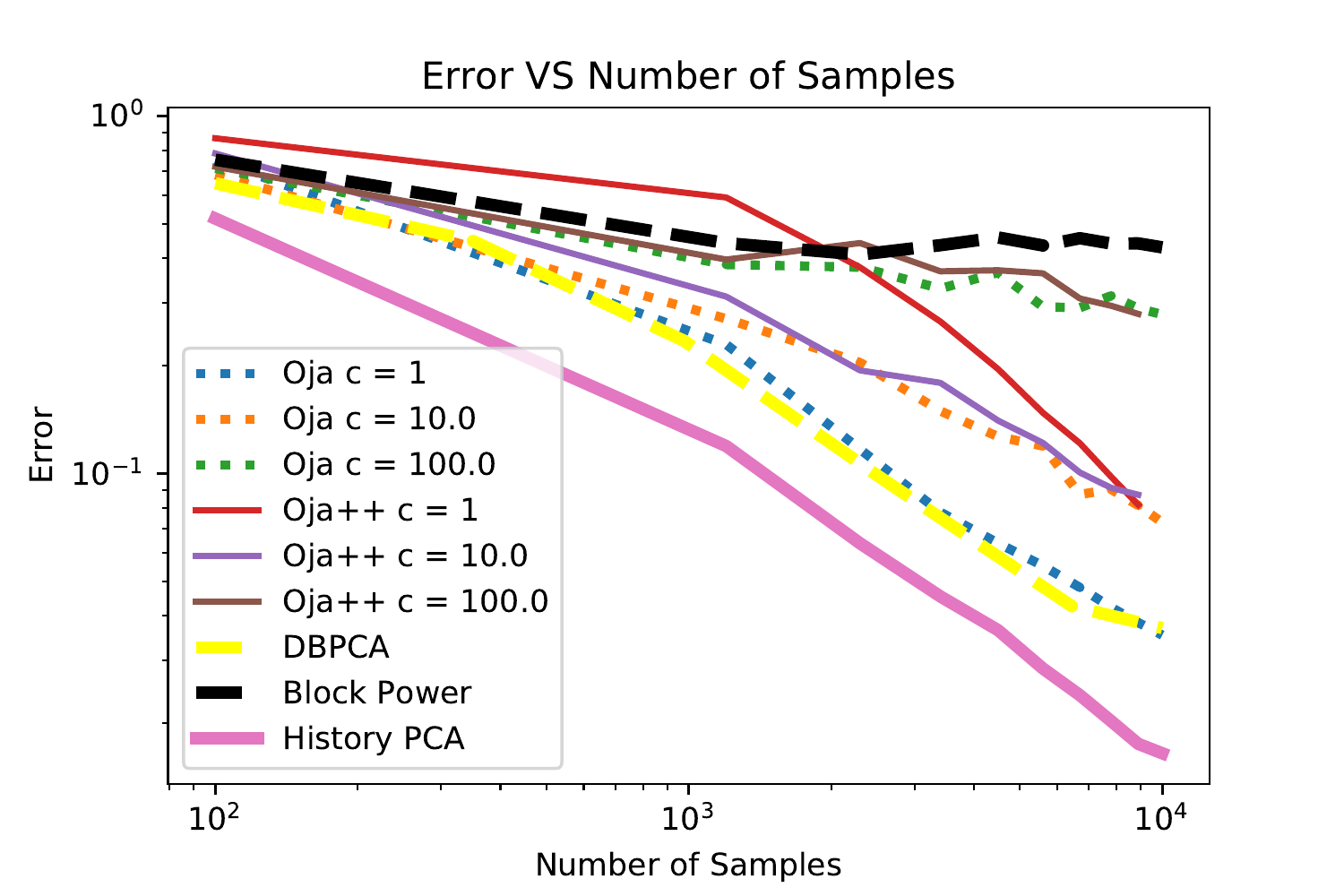}
    \label{fig:meovspre}
    }& \hspace{-10pt}
    \subfloat[$d = 1000, B = 100, k = 5, sig = 0.8$\label{fig:n10000_d1000_B100_k10_iter1}]{
    \includegraphics[width=0.32\linewidth]{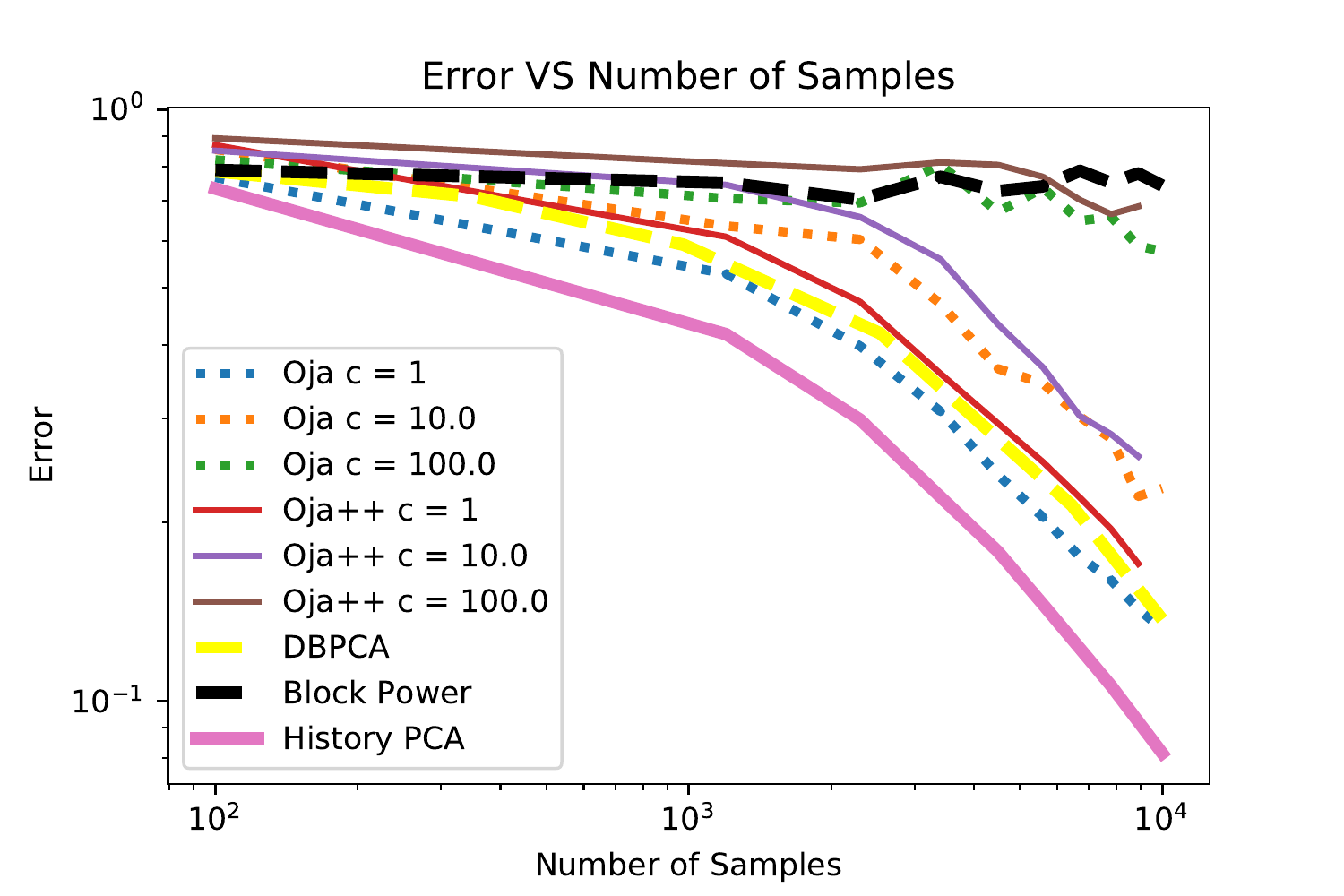}    \label{fig:timevspre}
    }\\
       \hspace{-10pt}
    \subfloat[$d = 1000, B = 100, k = 10, sig = 0.1$ \label{fig:n10000_d1000_B100_k1_iter1}]{
    \includegraphics[width=0.32\linewidth]{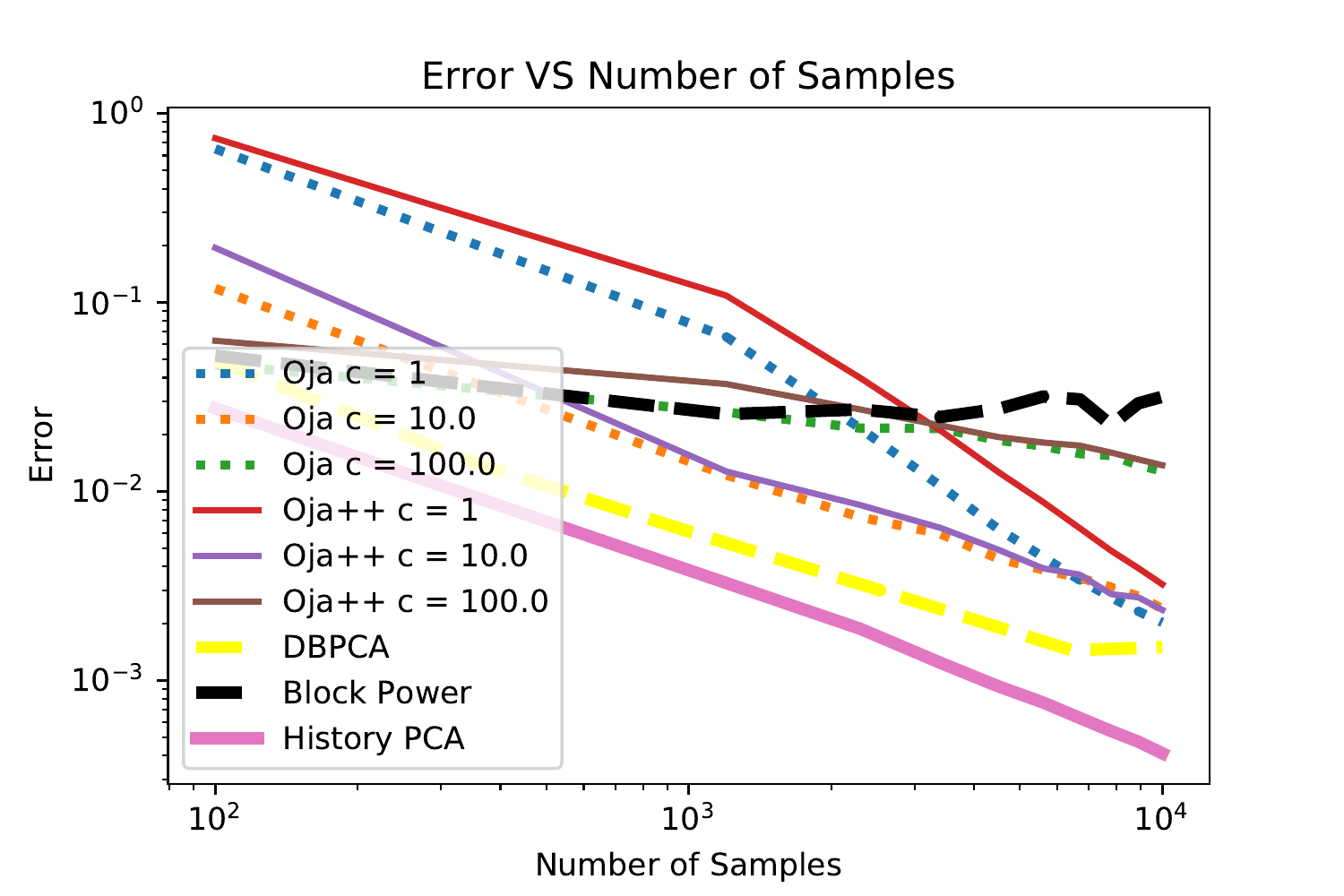}
    \label{fig:meovspre}
    } & \hspace{-10pt}
    \subfloat[$d = 1000, B = 100, k = 10, sig = 0.5$ \label{fig:n10000_d1000_B100_k5_iter1}]{
    \includegraphics[width=0.32\linewidth]{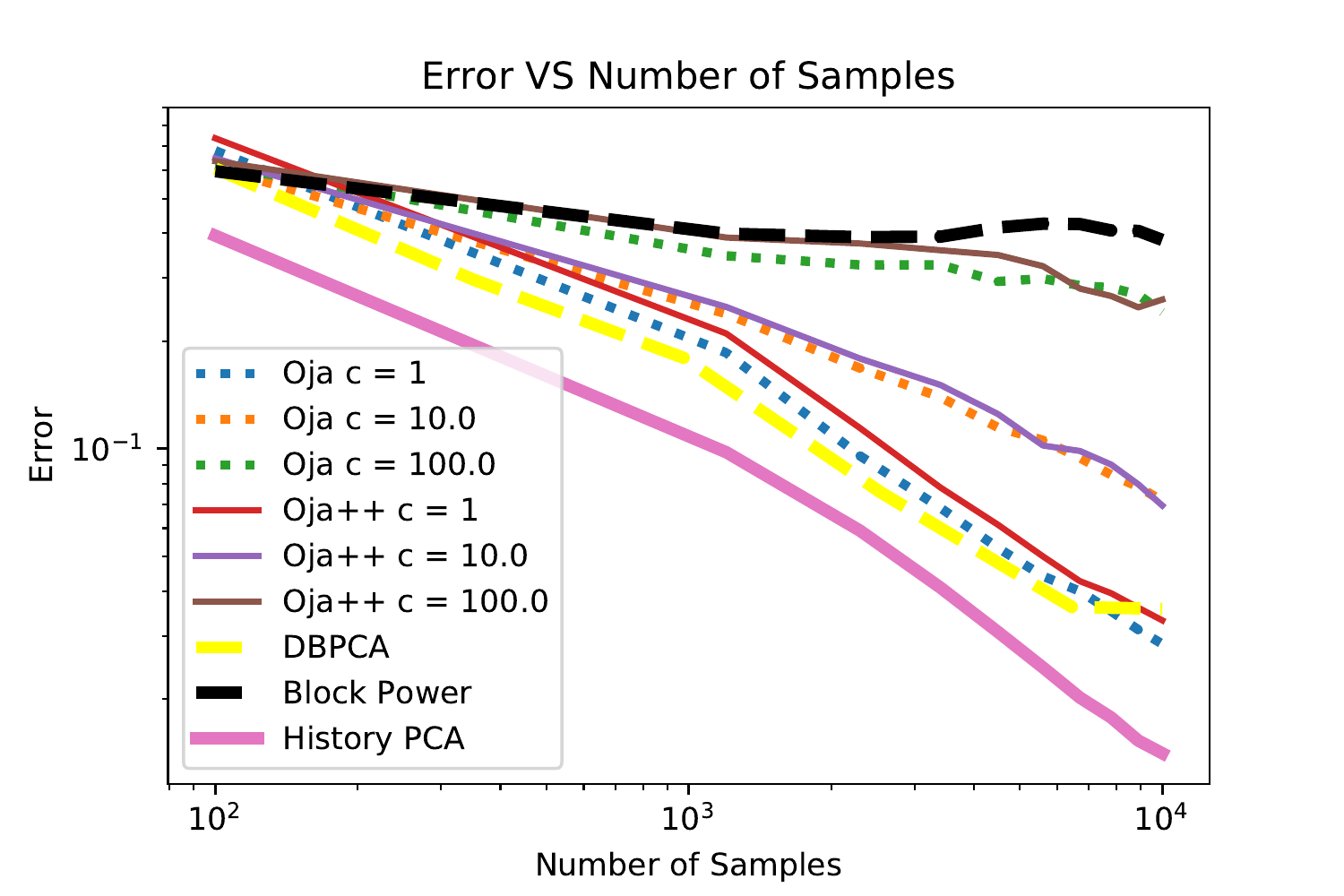}
    \label{fig:meovspre}
    }& \hspace{-10pt}
    \subfloat[$d = 1000, B = 100, k = 10, sig = 0.8$\label{fig:n10000_d1000_B100_k10_iter1}]{
    \includegraphics[width=0.32\linewidth]{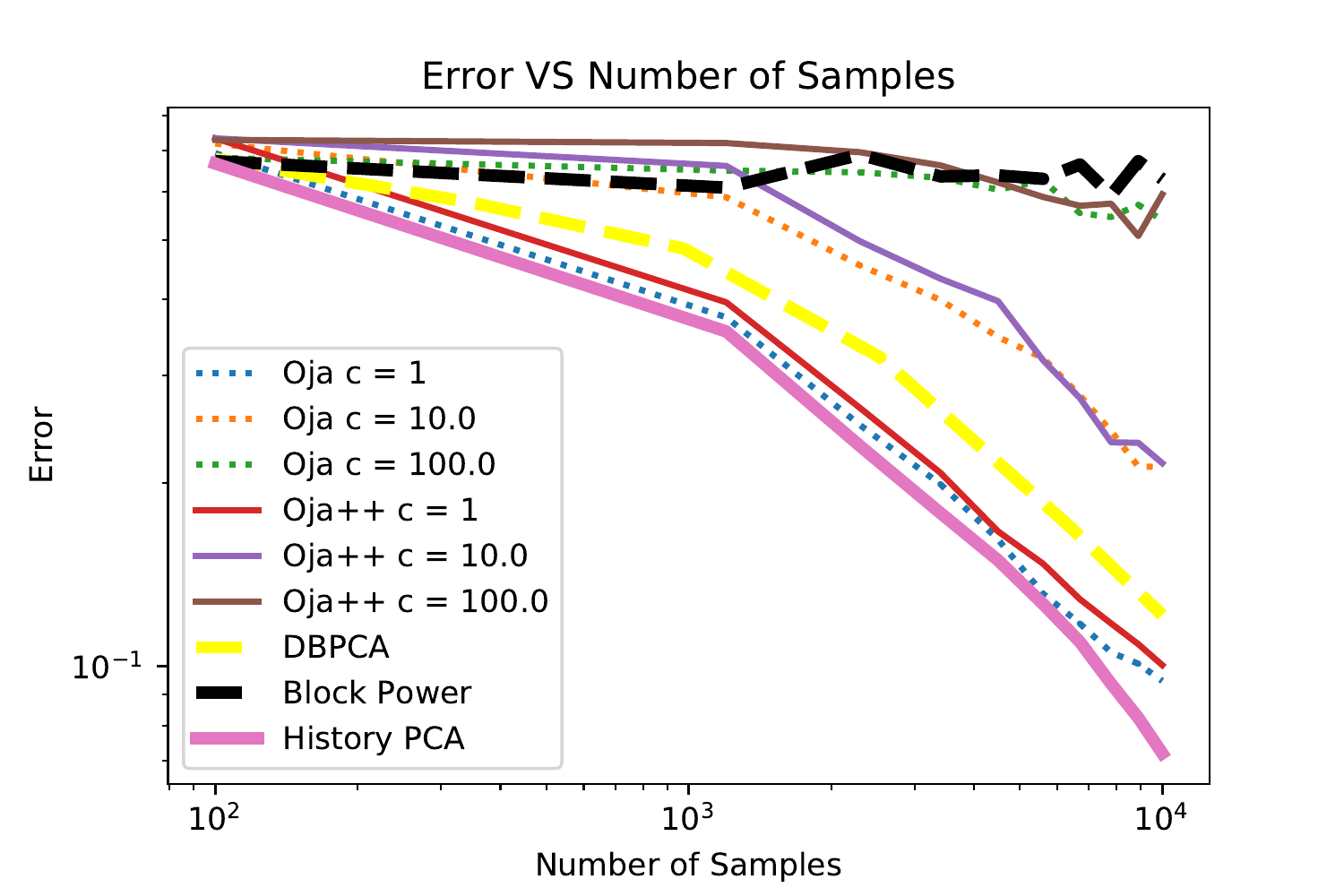}    \label{fig:timevspre}
    }
  \end{tabular}
  \caption{Comparison of streaming PCA algorithms on simulated data sets (d = 1000).} 
  \label{fig: simulated_data}
\end{figure*}

We set Oja's algorithm and Oja$^{++}$ algorithm with step sizes $\frac{c}{t}$, where $t$ is the number of iterations and $c$ is the tuning parameter.  We obtain the results for $c= 10^j, j= -6, -5, -4, -3, -2, -1, 0, 1, 2, 3, 4, $
but only include the best three $c$ values in Figure~\ref{fig: simulated_data2} and Figure~\ref{fig: simulated_data}.
The same experimental setting in \cite{IM13a} is used in order to compare our methods with theirs. Here we tried $n = 10000$ and varied the noise standard deviation $\sigma = 0.1, 0.5, 0.8$ for the simulated data sets. 
Thus, $\Sigma = UU^\top +\sigma^2 I$, where $U$ is the $d \times k$ true orthonormal matrix we are trying to recover. 

We consider two scenarios:  $d = 100$ and $d = 1000$. For each scenario, we try to recover the top $k = 1, 5, 10$ eigenvectors of the data sets respectively. We also vary the block size $B = 10, 100$. The performances of the algorithms are  evaluated based on the largest-principal-angle metric, which measures the distances between the estimated principal vectors and the ground truth. 


Figure~\ref{fig: simulated_data2} shows the results for $d = 100$ and Figure~\ref{fig: simulated_data} shows the results for $d = 1000$. Both figures suggest that the History PCA method is the best for the case of recovering the top eigenvector, as well as for cases  recovering  additional top eigenvectors, regardless of the block size,  noise level, and whether $d$ is large or small.   
What's more, the error from the History PCA method continues to decrease as the sample size of streaming data increases,  while some other streaming methods stop improving at a very early stage. 
From the simulations, we conclude that the performance of the block power method depends highly on the block size $B$. The performance of the block power method improves as $B$ increases. But the History PCA method does not depend on the block size, which is an advantage over the block power method. 
DBPCA's performance is sometimes much better than the block power method and outperforms the Oja's algorithm and Oja$^{++}$ algorithm in most scenarios. But it is less stable and less accurate than our History PCA method. 
The performance of Oja's algorithm and Oja$^{++}$ algorithm depends highly on the tuning of step size. There is no universal good step size for Oja's algorithm and Oja$^{++}$ algorithm. The History PCA method does not need to tune the step size but yields superior performance than those of the best tuned Oja's algorithm and Oja$^{++}$ algorithm in all scenarios. 

\subsection{Large-scale real data}

\begin{figure*}[h]
\vspace*{-0.1in}
  \centering
  \begin{tabular}{ccc}
    \subfloat[NIPS, $B=10$, $k=1$]{
    \includegraphics[width=0.32\linewidth]{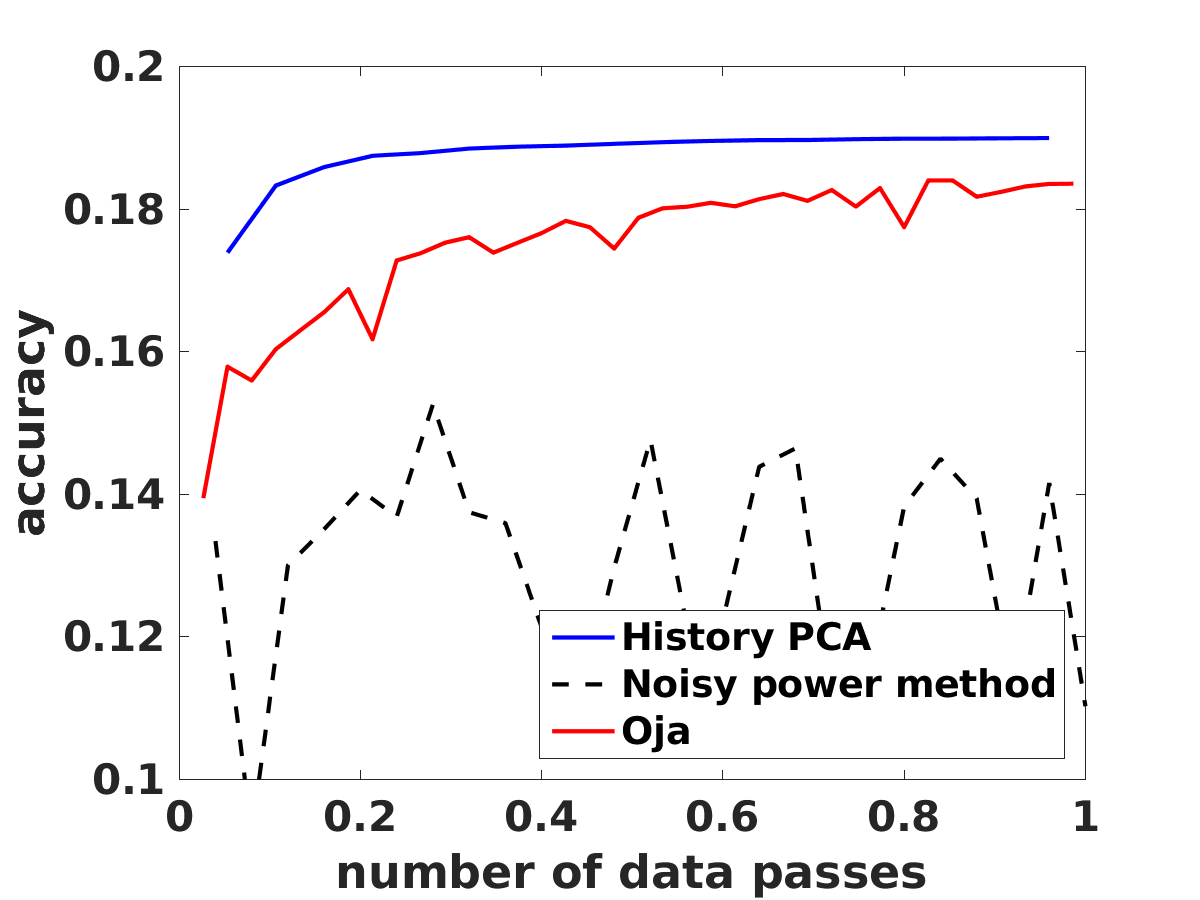}
    \label{fig:compare}
    }& \hspace{-10pt}
    \subfloat[NIPS, $B=100$, $k=1$]{
    \includegraphics[width=0.32\linewidth]{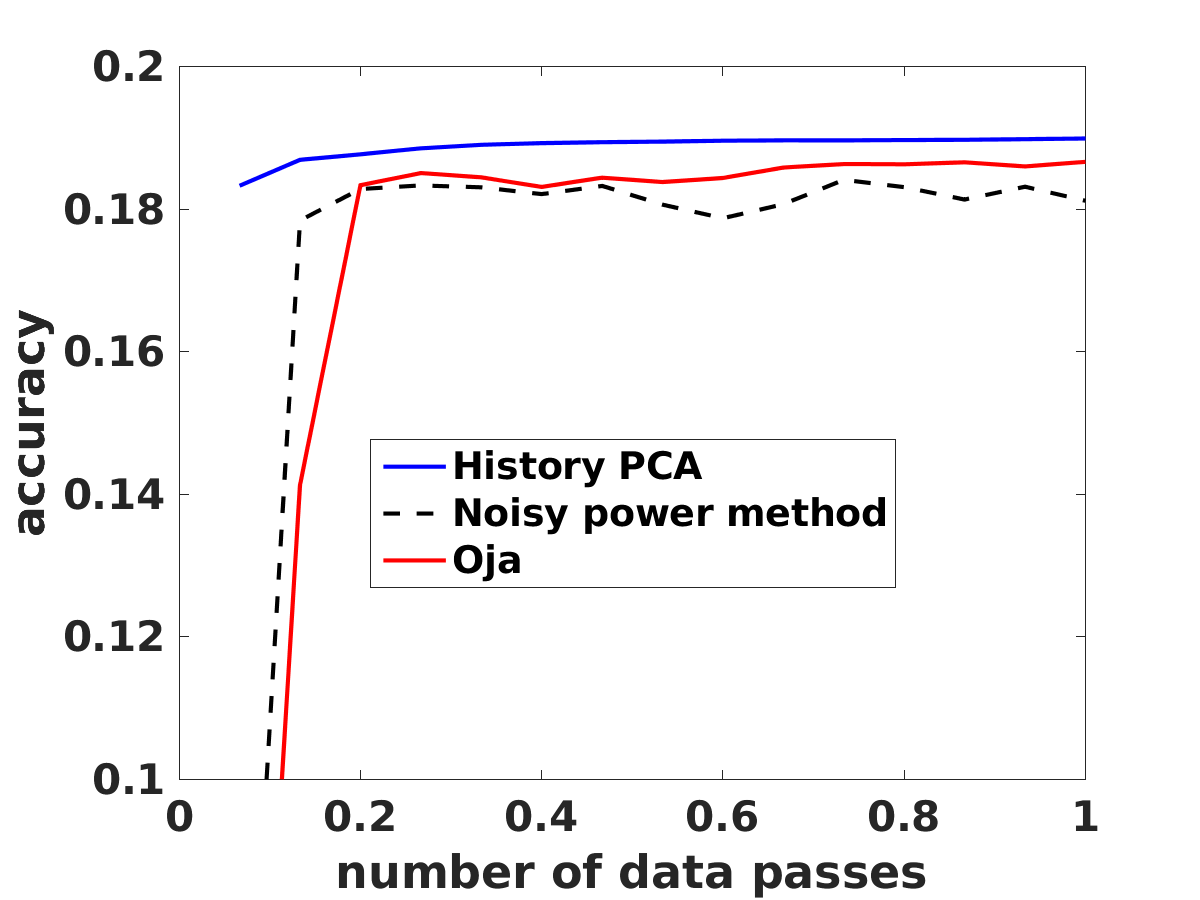}
    \label{fig:timevspre}
    }
  & \hspace{-10pt}
    \subfloat[NIPS, $B=100$, $k=10$]{
    \includegraphics[width=0.32\linewidth]{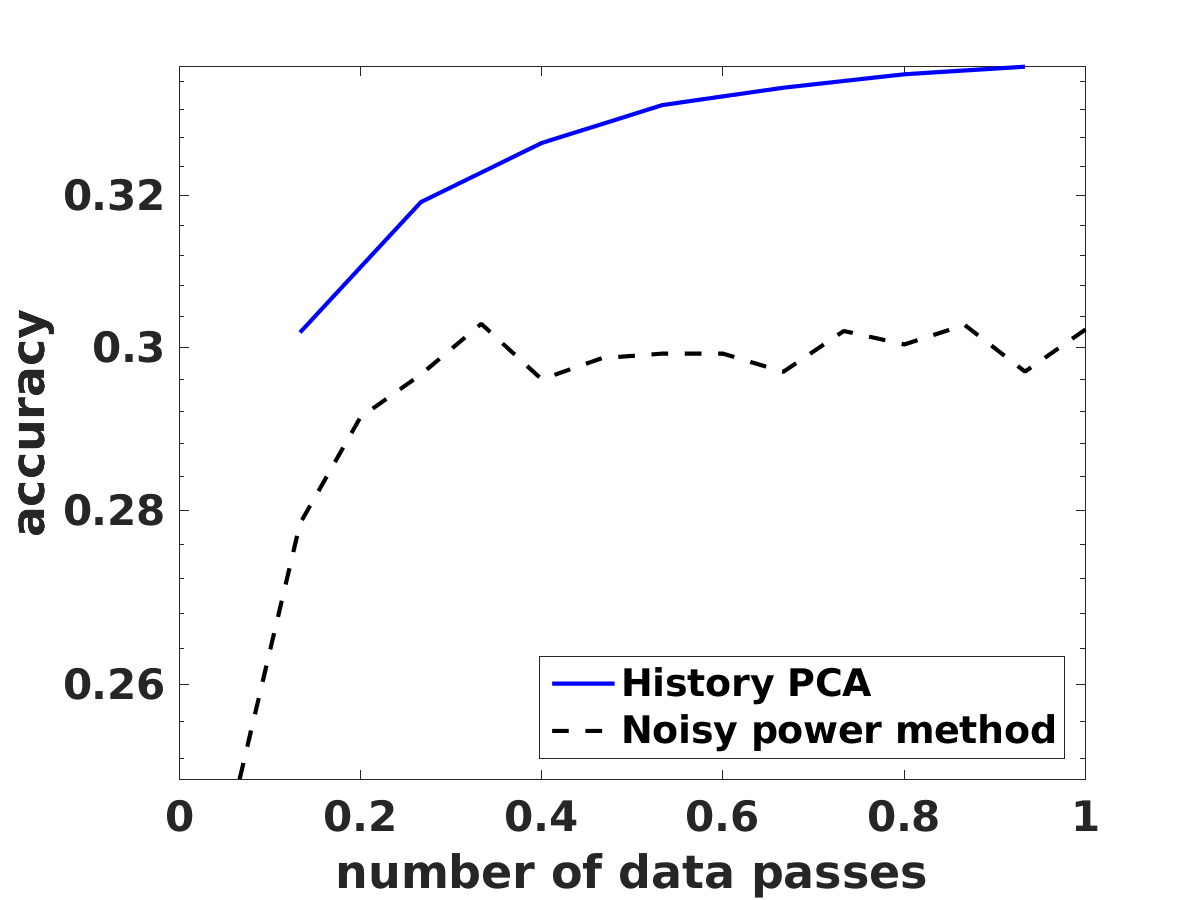}
    \label{fig:meovspre}
    }\\\vspace*{-0.1in}
       \subfloat[NYTimes, $B=10$, $k=1$]{
    \includegraphics[width=0.32\linewidth]{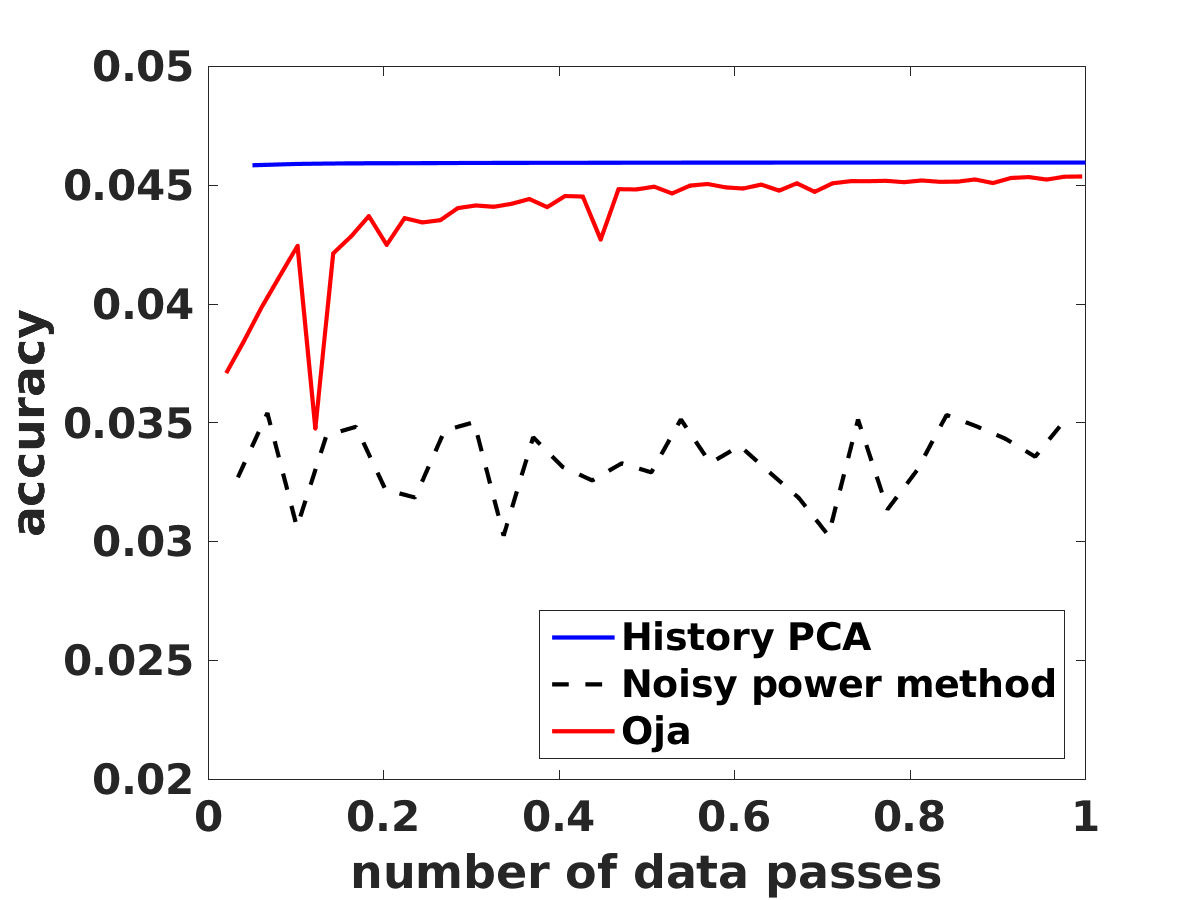}
    \label{fig:compare}
    }& \hspace{-10pt}
    \subfloat[NYTimes, $B=100$, $k=1$]{
    \includegraphics[width=0.32\linewidth]{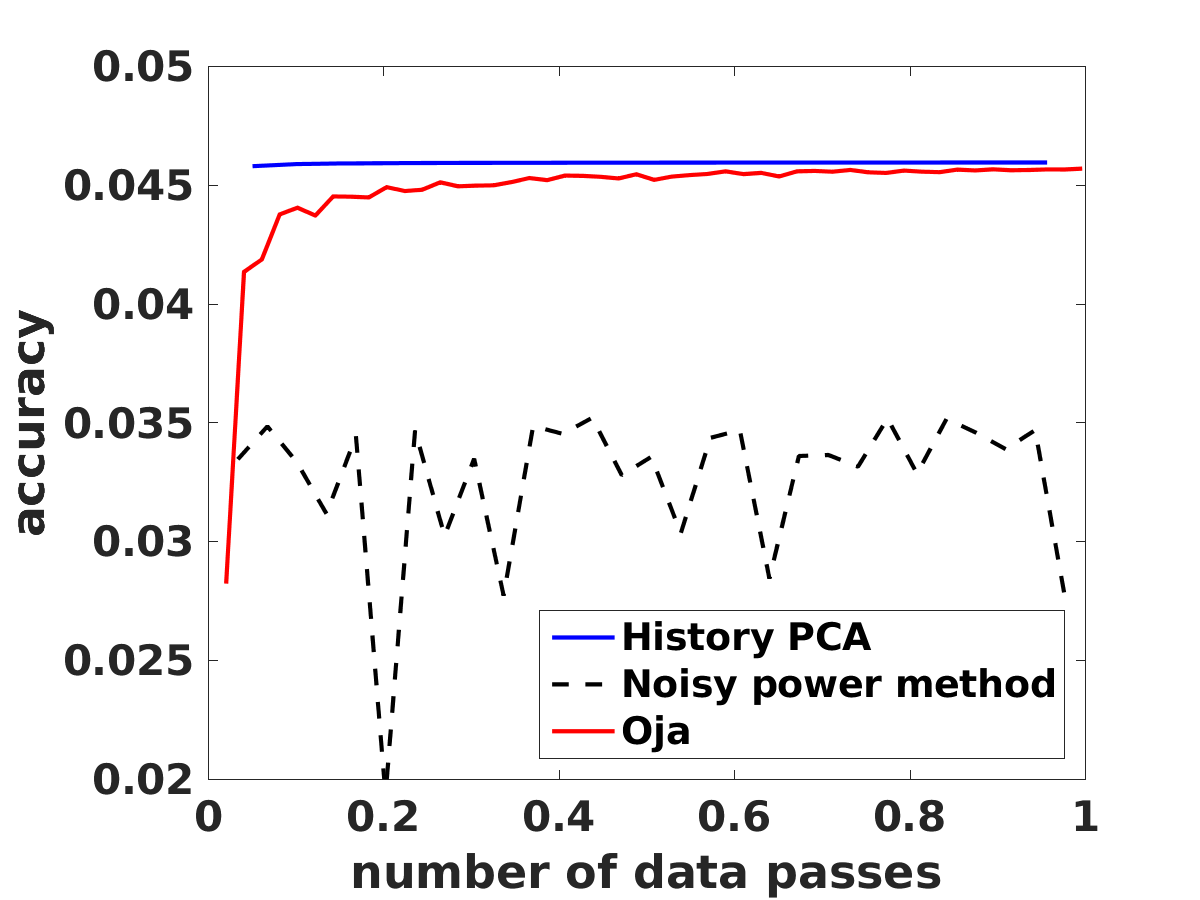}
    \label{fig:timevspre}
    }
  & \hspace{-10pt}
    \subfloat[NYTimes, $B=100$, $k=10$]{
    \includegraphics[width=0.32\linewidth]{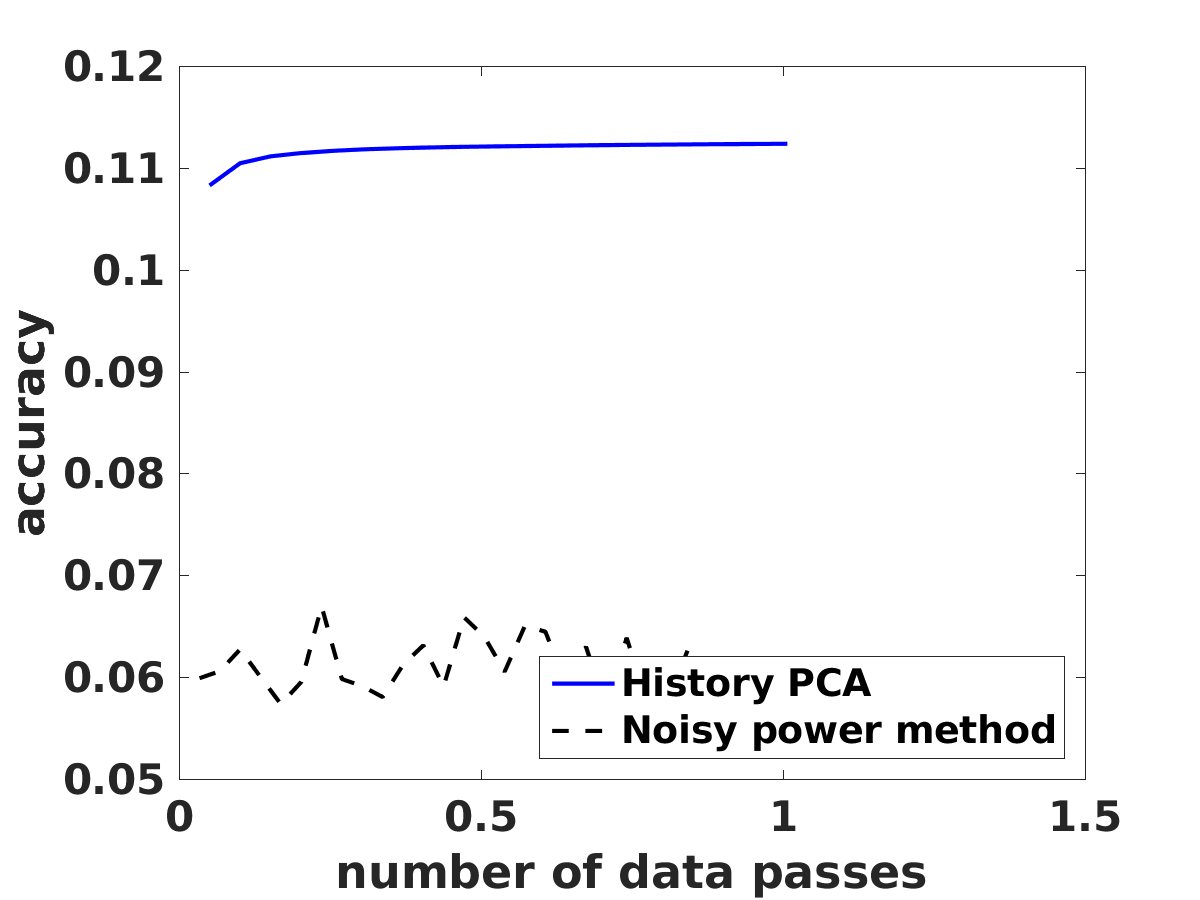}
    \label{fig:meovspre}
    }
  \end{tabular}
  \caption{Comparison of streaming PCA algorithms on real data sets (streaming setting). } 
  \label{fig:real_streaming}
\end{figure*}

\begin{figure*}[h]
\vspace*{-0.1in}
  \centering
  \begin{tabular}{cc}
    \subfloat[RCV1 $k=1$, data passes \label{fig:rcv1_iter}]{
    \includegraphics[width=0.35\linewidth]{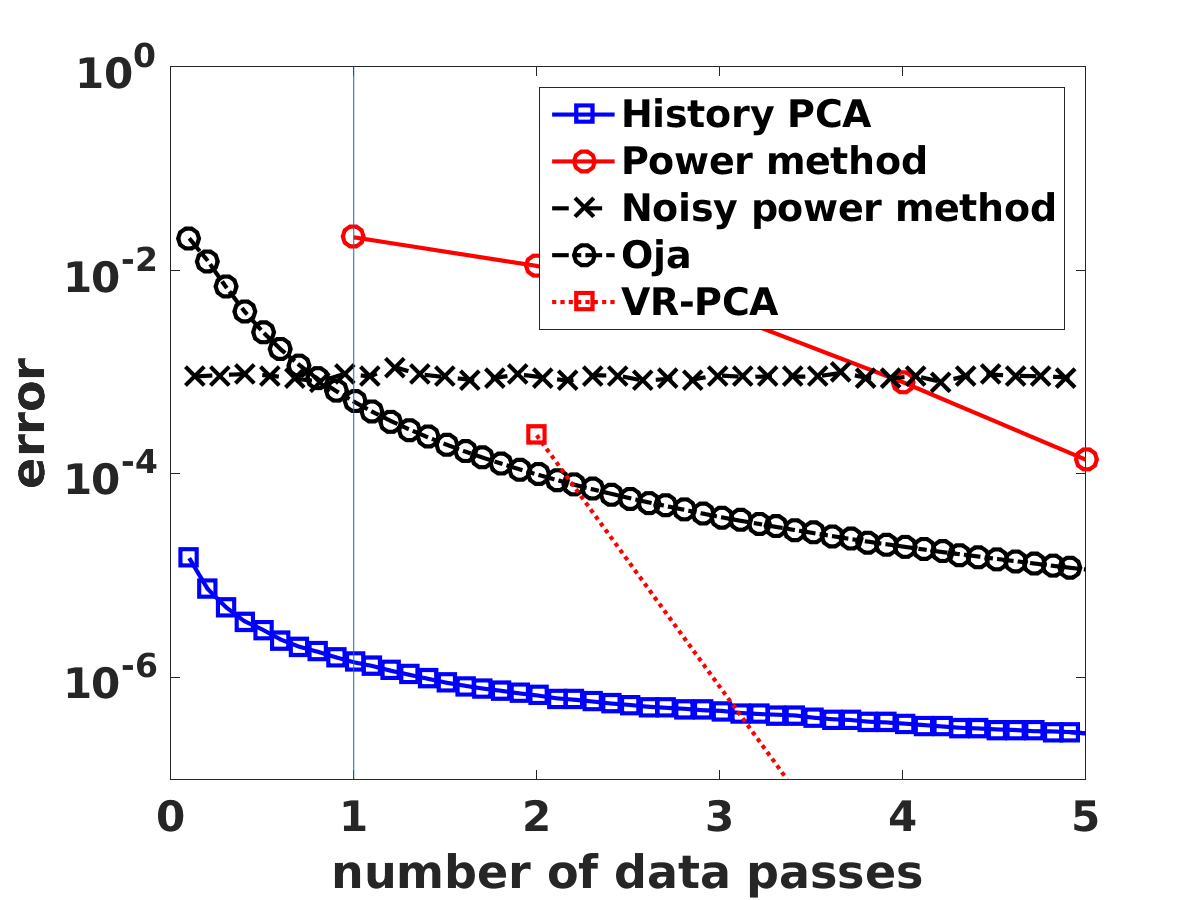}
    \label{fig:compare}
    }& \hspace{-10pt}
    \subfloat[KDDB $k=1$, data passes \label{fig:kddb_iter}]{
    \includegraphics[width=0.35\linewidth]{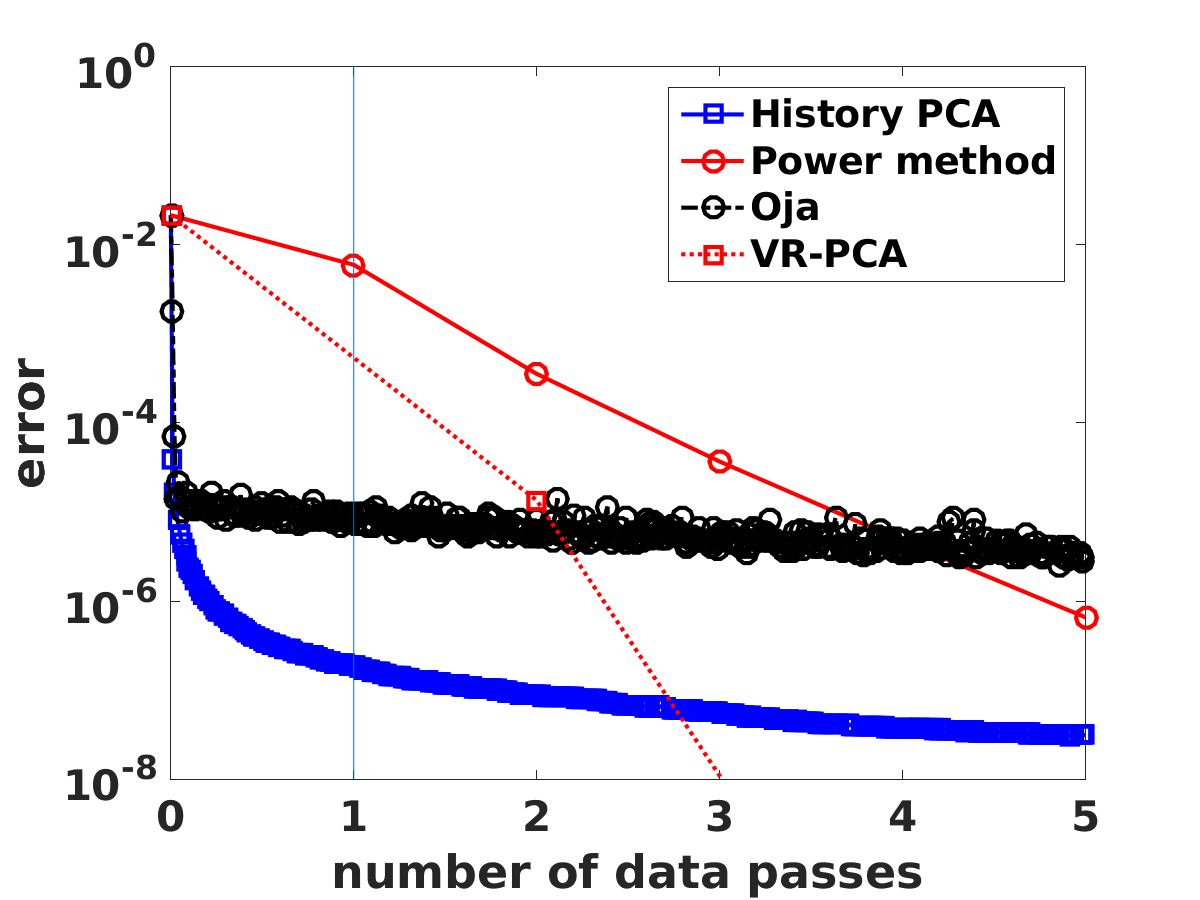}
    \label{fig:timevspre}
    } \\\vspace*{-0.1in}
    \subfloat[RCV1 $k=1$, time \label{fig:rcv1_time}]{
    \includegraphics[width=0.35\linewidth]{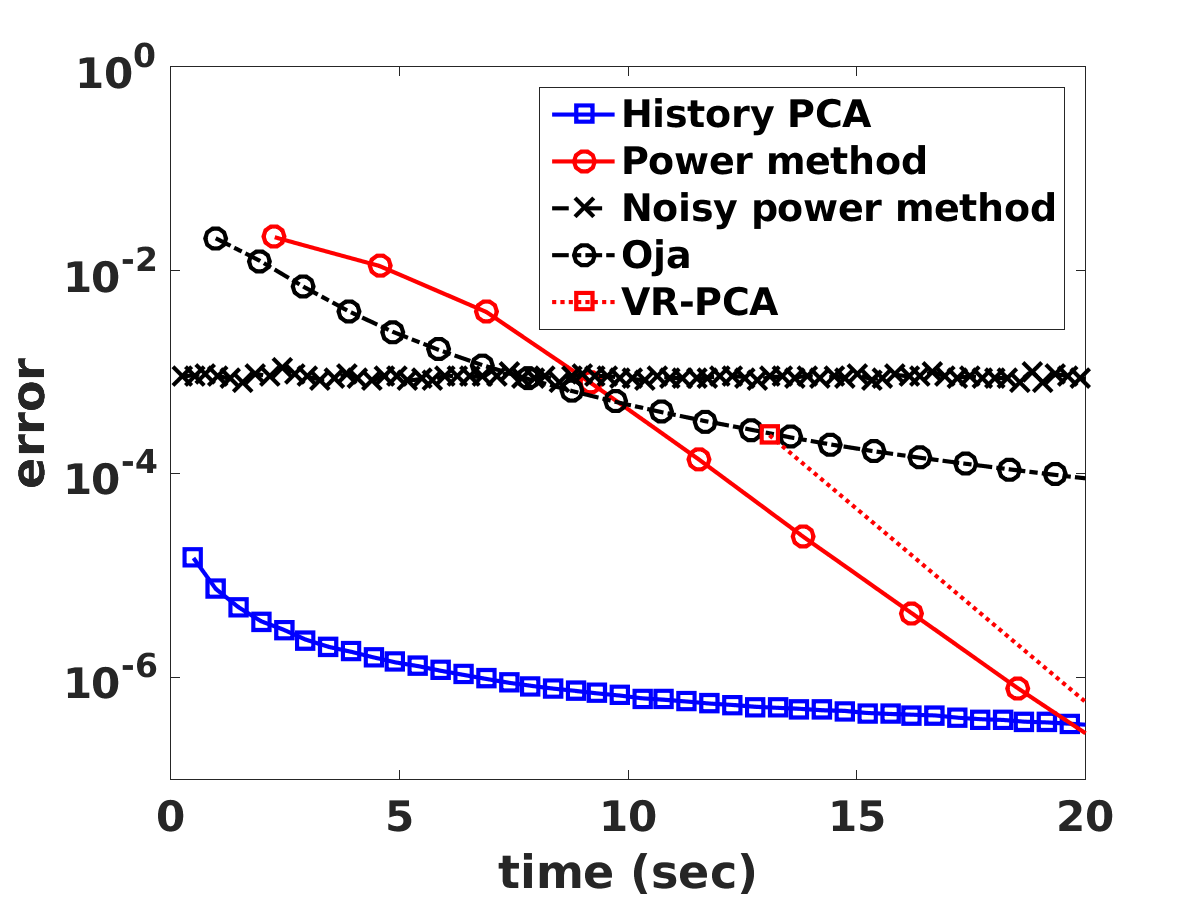}
    \label{fig:meovspre}
    } & \hspace{-10pt}
    \subfloat[KDDB $k=1$, time \label{fig:kddb_time}]{
    \includegraphics[width=0.35\linewidth]{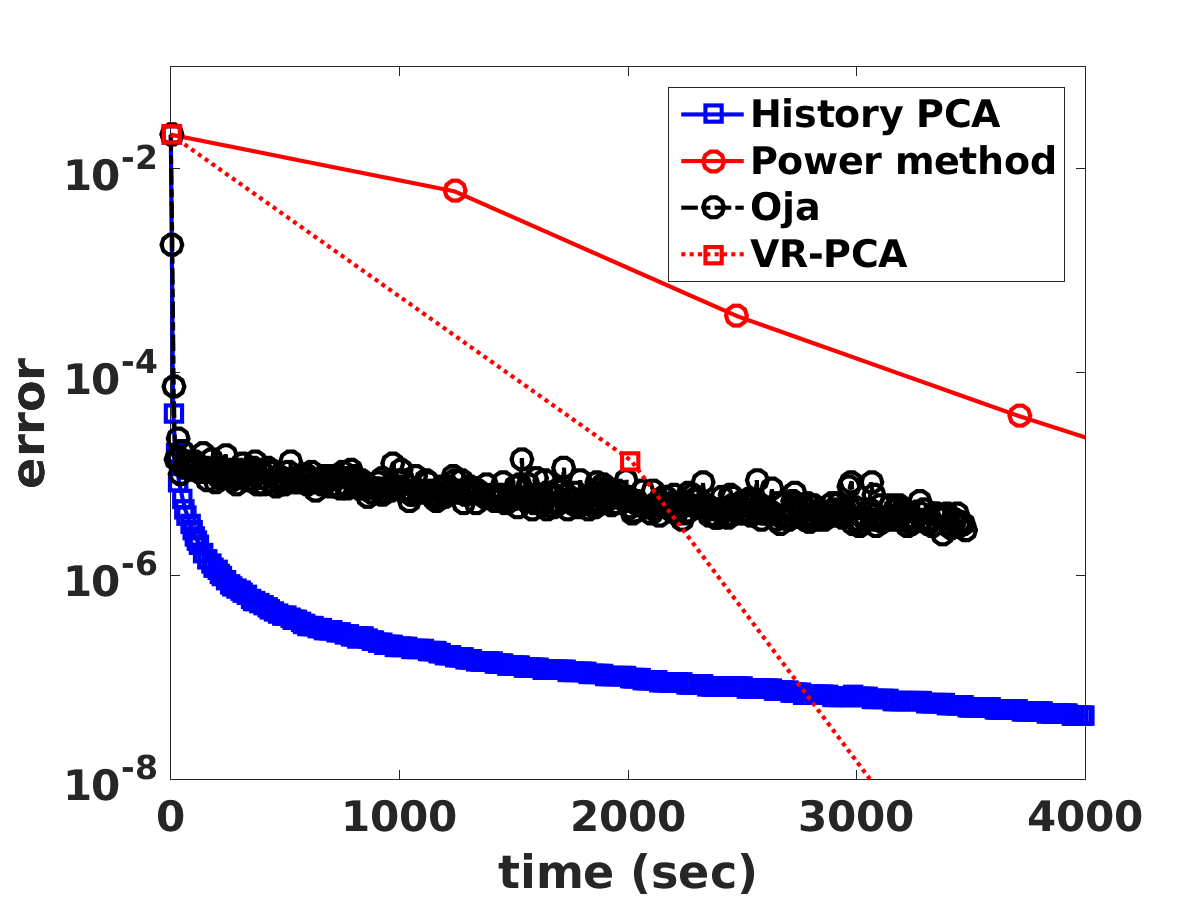}
    \label{fig:meovspre}
    }
  \end{tabular}
  \caption{Comparison of all batch and streaming PCA algorithms on large data sets with respect to number of data access (a, b) and run time (c, d).}
  \label{fig:real_large}
\end{figure*}

\begin{table}[h]
    \centering
    \begin{tabular}{cccc}
         &  \# samples & \# features & \# nonzeroes\\
         \hline
    NIPS  &       1500           &          12419           &   746,316                  \\
    \hline
    NYTimes &      300,000           &        102,660              &        69,679,427           \\
    \hline
    RCV1  & 677,399     &      47,236   &    49,556,258   \\
    \hline
    KDDB  & 19,264,097  & 29,890,095  &  566,345,888    \\
    \end{tabular}
    \caption{Data set statistics}
    \label{tab:datasets}
\end{table}
We compare our algorithm with Oja's algorithm, VR-PCA and noisy power method~\cite{IM13a, MH14a} 
on four real data sets. We do not include the Oja$^{++}$ algorithm as we observe that its performance is almost identical to the Oja's algorithm. NIPS and NYTimes data sets are downloaded from UCI data.
RCV1 and KDDB data sets are downloaded from LIBSVM data sets. A summary of our data sets is provided in Table~\ref{tab:datasets}.

We first consider the streaming setting, where the algorithms 
can only go through the data once. 
For a fair comparison, we set Oja's algorithm with step sizes $\frac{c}{t}$, where
$t$ is number of iterations and  $c$ ranges from $10^{-6}$ to $10^{4}$ 
in all the experiments. The best result of the Oja's algorithm is presented in the plots. 
We also compare the algorithms with different 
block size $B$ and at different target rank $k$. 
In practice we find Oja's algorithm ($k>1$) often numerically unstable and cannot achieve the comparable performance with other approaches, 
so we do not include their result for $k>1$. To evaluate the results, we use the widely used metric ``explained variance'', 
\begin{equation*}
    \frac{\text{trace}(W^\top  X^\top  X W)} { \|X\|_F^2},
\end{equation*}
to measure the quality of the solution (where $W$ is the computed solution).  
Note that the value is always less or equal to 1, and 
will be maximized when $W$ is the leading eigenvectors of $X^\top X$. 
In Figure~\ref{fig:real_streaming}, we observe that
our algorithm (History PCA) is very stable and consistently better 
in all settings (big or small $B$, big or small $k$). 

\subsection{Comparing streaming with non-streaming algorithms} 
In the final set of experiments, we go beyond the streaming setting and allow several passes of 
data. We want to test (1) In addition to having fewer number of data access, can our
algorithm also have short running time in practice? (2) If we go beyond the first pass of the data, 
can our algorithm keep on improving the solution by making more passes? 
(3) In practice, if there is a large data set with millions or billions of data points and we want to compute PCA,  should we use our algorithm instead of other batch PCA algorithms? 

In order to answer these questions, we test our algorithm on two large data sets: RCV1 and KDDB (with 20 million samples and 30 million features). 
Also, we add two strong batch algorithms into comparison, including power method and VR-PCA~\cite{OS15a}, 
which are the state-of-the-art PCA solvers. In order to clearly show the convergence speed, we compute the ``error'' by the largest eigenvalue minus the unnormalized explained variance $\bw^\top  X^\top  X \bw$, so ideally the error will converge to 0 (we are doing this in order to show the error in log-scale in the y-axis). 
The results are presented in Figure~\ref{fig:real_large}. In Figure~\ref{fig:rcv1_iter} and \ref{fig:kddb_iter}, we show the number of data passes versus errors, and the vertical line means the end
of the first data pass. We observe that even within the first data pass, the error of our algorithm can be bounded by $10^{-6}$ and after the first pass our algorithm is still able to improve the solution. We also observe that VR-PCA achieves better performance with more passes of data and sufficient time while we argue that an error of order $10^{-6}$ is accurate enough for practical use. 


Finally we compare the real run time with all the other algorithms. In Figure~\ref{fig:rcv1_time}, 
since the data set is not that large, we store the data set in memory and perform in-memory
data access, while for KDDB data (Figure~\ref{fig:kddb_time}), since the data size is larger than memory, we load data block by block and compare the overall running time.
The results show that our algorithm is still 
much better than other methods in terms of run time.
The only real competitor, again, is VR-PCA, which will catch up our algorithm
when reaching an error of order $10^{-7}$. 
In the future, 
it will be interesting to apply a variance reduction technique to speed up the convergence of our algorithm after the first pass of data. 

\section{Conclusions}
We propose History PCA, a hyperparameter-free streaming PCA algorithm which utilizes past information effectively.
We extend our algorithm to solve the rank-$k$ PCA problems with $k>1$. What's more, we provide the theoretical guarantees and convergence rate for our algorithm, and we show that the convergence speed is faster than existing algorithms on both synthetic and real data sets. 


\bibliography{pca}  
\bibliographystyle{ACM-Reference-Format}
\appendix

\section{Proof of Lemma 2}
\begin{proof}
Since $B_n = (I +\eta_n A_n) \cdots (I +\eta_1 A_1)$ with $B_0 = I$ and $B_{n+n_0} = C_{n, n_0} B_{n_0}$, we know that $C_{n, n_0} = (I +\eta_{n_0+n} A_{n_0+n}) \cdots (I +\eta_{n_0+1} A_{n_0+1}).$
By Lemma 6 of \cite{PJ16a}, we know that with probability at least $1 - \delta$,
\begin{align*}
\sin^2(\bv, \frac{B_{n+n_0}\bw}{\|B_{n+n_0}\bw\|_2}) 
&= 1-(\frac{\bv^\top B_{n+n_0}\bw}{ \|B_{n+n_0}\bw\|_2})^2  \\
& \leq \frac{c_1 \log (1 / \delta)}{\delta} \frac{Tr(V_\perp^\top B_{n+n_0} B_{n+n_0}^\top V_\perp)}{\bv^\top B_{n+n_0} B_{n+n_0}^\top \bv}  \\
& \leq \frac{c_1 \log (1 / \delta)}{\delta} \frac{Tr(V_\perp^\top C_{n, n_0} B_{n_0}  B_{n_0}^\top C_{n, n_0}^\top V_\perp)}{\bv^\top C_{n, n_0} B_{n_0} B_{n_0}^\top C_{n, n_0}^\top \bv},
\end{align*}
where $c_1$ is an absolute constant. 
Since 
\begin{align*}
 Tr(V_\perp^\top C_{n, n_0} B_{n_0}  B_{n_0}^\top C_{n, n_0}^\top V_\perp) 
&= Tr(B_{n_0}  B_{n_0}^\top C_{n, n_0}^\top V_\perp V_\perp^\top C_{n, n_0} ) \\
& \leq \| B_{n_0}  B_{n_0}^\top \|_2 Tr(C_{n, n_0}^\top V_\perp V_\perp^\top C_{n, n_0} ),  
\end{align*}
\begin{align*}
 \bv^\top C_{n, n_0} B_{n_0} B_{n_0}^\top C_{n, n_0}^\top \bv 
&= Tr(B_{n_0}  B_{n_0}^\top C_{n, n_0}^\top \bv \bv^\top C_{n, n_0} )  \\
& \geq \sigma_n( B_{n_0}  B_{n_0}^\top ) Tr(C_{n, n_0}^\top \bv \bv^\top C_{n, n_0} ) \\
& \geq \sigma_n( B_{n_0}  B_{n_0}^\top )  \bv^\top C_{n, n_0} C_{n, n_0}^\top \bv,  
\end{align*}
we have 
\begin{align*}
\sin^2(\bv, \frac{B_{n+n_0}\bw}{\|B_{n+n_0}\bw\|_2}) 
& \leq \frac{c_1 \log (1 / \delta)}{\delta} \frac{Tr(V_\perp^\top C_{n, n_0} B_{n_0}  B_{n_0}^\top C_{n, n_0}^\top V_\perp)}{\bv^\top C_{n, n_0} B_{n_0} B_{n_0}^\top C_{n, n_0}^\top \bv} \\
& \leq \frac{c_1 \log (1 / \delta)}{\delta} \frac{\| B_{n_0}  B_{n_0}^\top \|_2 Tr(C_{n, n_0}^\top V_\perp V_\perp^\top C_{n, n_0} ) 
}{ \sigma_n( B_{n_0}  B_{n_0}^\top )  \bv^\top C_{n, n_0} C_{n, n_0}^\top \bv  } \\
& \leq \frac{c_2 \log (1 / \delta)}{\delta} \frac{  Tr(C_{n, n_0}^\top V_\perp V_\perp^\top C_{n, n_0} ) }{  \bv^\top C_{n, n_0} C_{n, n_0}^\top \bv  } \\
& \leq \frac{c_2 \log (1 / \delta)}{\delta} \frac{  Tr( V_\perp^\top C_{n, n_0} C_{n, n_0}^\top V_\perp) }{  \bv^\top C_{n, n_0} C_{n, n_0}^\top \bv  }, 
\end{align*}
where $c_2$ is an absolute constant.
\end{proof}

\section{Proof of Theorem 3}

\begin{proof}
Define the step sizes $\hat{\eta}_t = \eta_{t+n_0} = \frac{1}{n_0 + t}$ for all positive integer $t$. Then we have $C_{n, n_0} = (I +\hat{\eta}_n A_{n_0+n}) \cdots (I +\hat{\eta}_1 A_{n_0+1}).$
With $n_0 >  \max( 4 \mathcal{M}, 18 \frac{\nu +\lambda_1^2}{\log (1 + \frac{\delta}{100})})$, we have $\hat{\eta}_t = \frac{1}{n_0 + t} \leq \frac{1}{4 \max(\mathcal{M}, \lambda_1)}.$ Thus, $\hat{\eta}_t$ satisfy the conditions in Theorem 3.1 of \cite{PJ16a}. Define $\bar{ \mathcal{V}} = \mathcal{V} + \lambda_1^2$. So, we have a bound for the error 
\begin{align*}
1 - (\bw_{n +n_0}^T v_1)^2 
& \leq \frac{c_1 \log (1 / \delta)}{\delta} \frac{  Tr(C_{n, n_0}^\top V_\perp V_\perp^\top C_{n, n_0} ) 
}{  v^\top C_{n, n_0} C_{n, n_0}^\top v   } \\
&\leq \frac{1}{Q} \exp (5 \bar{ \mathcal{V}} \sum_{i = 1}^n \hat{\eta}_i^2) (d \exp ( - 2(\lambda_1 - \lambda_2) \sum_{i = 1}^n\hat{\eta}_i) \\
&\ \ \ \ + \mathcal{V} \sum_{i = 1}^n  \hat{\eta}_i^2 \exp ( -\sum_{j = i + 1}^n 2 \hat{\eta}_j (\lambda_1 - \lambda_2))),  
\end{align*}
where $Q = \frac{\delta^2}{c_1 \log(1/\delta)} ( 1 -\frac{1}{\sqrt{\delta}} \sqrt{ \exp (18 \bar{ \mathcal{V}} \sum_{i = 1}^n \hat{\eta}_i^2) - 1} )$.

Since $\hat{\eta}_t = \frac{1}{n_0 + t}$, we have 
\begin{align*}
\sum_{i = 1}^{n}{\hat{\eta}_t^2} &\leq \frac{1}{n_0}  \\
\sum_{i = 1}^{n}{\hat{\eta}_t^2} &\leq \frac{\log (1 + \frac{\delta}{100})}{18 \bar{\mathcal{V}} }\\
\exp (18 \bar{\mathcal{V}} \sum_{i = 1}^{n}{\hat{\eta}_t^2}) - 1&\leq \frac{\delta}{100} \\
1 -\frac{1}{\sqrt{\delta}} \sqrt{ \exp (18 \bar{ \mathcal{V}} \sum_{i = 1}^n \hat{\eta}_i^2) - 1} & > \frac{9}{10}. 
\end{align*}
Thus, we can conclude that $Q \geq \frac{9}{10} \frac{\delta^2}{c_1 \log(1/\delta)}$. 

Besides, we have $\exp (5 \bar{ \mathcal{V}} \sum_{i = 1}^n \hat{\eta}_i^2) \leq 1+\frac{\delta}{100}.$
Therefore, 
\begin{align*}
\frac{\exp (5 \bar{ \mathcal{V}} \sum_{i = 1}^n \hat{\eta}_i^2)}{Q} & \leq (1+\frac{\delta}{100})\frac{10}{9} \frac{c_1 \log(1/\delta)}{\delta^2} \leq \frac{c \log(1/\delta)}{\delta^2}. 
\end{align*}
for some constant $c$. Since $\sum_{t = 1}^n \hat{\eta}_t \geq \log (1 + \frac{n}{n_0} )$, we have $$\exp (-2(\lambda_1 - \lambda_2) \sum_{t = 1}^n \hat{\eta}_t) \leq ( \frac{n_0}{n_0 + n} )^{2(\lambda_1 - \lambda_2)}.$$
Thus, 
\begin{align*}
&\ \ \ \ \sum_{i = 1}^n \hat{ \eta}_i^2 \exp ( - \sum_{j = i + 1}^n 2\hat{ \eta}_j (\lambda_1 - \lambda_2)) \\
& = \sum_{i = 1}^n  \frac{1}{(n_0 + i)^2} \exp ( -  2(\lambda_1 - \lambda_2) \sum_{j = i + 1}^n  \hat{\eta}_j) \\
&\leq \sum_{i = 1}^n  \frac{1}{(n_0 + i)^2} \exp (  2(\lambda_1 - \lambda_2) \log \frac{i + n_0 + 1}{n +n_0+1}) \\
&\leq \sum_{i = 1}^n  \frac{1}{(n_0 + i)^2}  ( \frac{i + n_0 + 1}{n +n_0+1})^{2(\lambda_1 - \lambda_2)} \\
&\leq \sum_{i = 1}^n  \frac{1}{(n_0 + i)^2} \frac{(n_0 + i + 1)^{2(\lambda_1 - \lambda_2)}}{(n_0 + i)^{2(\lambda_1 - \lambda_2)}}  ( \frac{i + n_0}{n +n_0+1})^{2(\lambda_1 - \lambda_2)} \\
&\leq  (\frac{n_0 + 2}{n_0 + 1})^{2(\lambda_1 - \lambda_2)} \sum_{i = 1}^n  \frac{1}{(n_0 + i)^2} ( \frac{i + n_0}{n +n_0+1})^{2(\lambda_1 - \lambda_2)} \\
&\leq  (\frac{n_0 + 2}{n_0 + 1})^{2(\lambda_1 - \lambda_2)} ( \frac{1}{n +n_0+1})^{2(\lambda_1 - \lambda_2)}\sum_{i = 1}^n  ( i + n_0)^{2(\lambda_1 - \lambda_2) - 2} 
\end{align*}


Since $\lambda_1 \neq \lambda_2$, we can always scale up the matrices $A_i$'s so that $2(\lambda_1 - \lambda_2) > 1$. Now we have 
\begin{align*}
&\sum_{i = 1}^n \hat{ \eta}_i^2 \exp ( - \sum_{j = i + 1}^n 2\hat{ \eta}_j (\lambda_1 - \lambda_2)) \\
&\leq  (\frac{n_0 + 2}{n_0 + 1})^{2(\lambda_1 - \lambda_2)} ( \frac{1}{n +n_0+1})^{2(\lambda_1 - \lambda_2)} \frac{(n + n_0)^{2(\lambda_1 - \lambda_2) - 1}}{2(\lambda_1 - \lambda_2) - 1}  \\
&\leq  (\frac{n_0 + 2}{n_0 + 1})^{2(\lambda_1 - \lambda_2)} \frac{1}{2(\lambda_1 - \lambda_2) - 1} \frac{1}{n +n_0}. 
\end{align*}

In conclusion, we have
\begin{align*}
1 - (\bw_{n +n_0}^T v_1)^2 & \leq \frac{c \log(1/\delta)}{\delta^2} (d  ( \frac{n_0}{n_0 + n} )^{2(\lambda_1 - \lambda_2)}  \\
& + (\frac{n_0 + 2}{n_0 + 1})^{2(\lambda_1 - \lambda_2)} \frac{1}{2(\lambda_1 - \lambda_2) - 1} \frac{1}{n +n_0} )\\
&= O(\frac{\mathcal{V}}{|2(\lambda_1 - \lambda_2) - 1|} \frac{1}{n +n_0}) 
\end{align*}

\end{proof}

\end{document}